\documentclass[manuscript]{acmart}
\AtBeginDocument{  \providecommand\BibTeX{{    \normalfont B\kern-0.5em{\scshape i\kern-0.25em b}\kern-0.8em\TeX}}}

\usepackage{url}   \usepackage{graphicx}   \usepackage{natbib}  \usepackage{caption}     \usepackage{algorithm}
\usepackage{algorithmic}
\usepackage{array}
\usepackage{arydshln}

\usepackage{newfloat}
\usepackage{listings}
\usepackage{hyperref}
\usepackage{amsmath}
\usepackage{nicefrac}
\usepackage{tikz}
\usetikzlibrary{automata,calc,fit,positioning}
\usepackage{wrapfig}
\usepackage{multirow}
\usepackage{xspace}
\usepackage{makecell}
\usepackage{longtable}
\usepackage{soul}
\usepackage{xspace}
\usepackage{pifont}
\usepackage{soul}
\usepackage{caption}
\usepackage{tabto}    
\usepackage[utf8]{inputenc} \usepackage[T1]{fontenc}    \usepackage{hyperref}       \usepackage{booktabs}       \usepackage{nicefrac}       \usepackage{microtype}      \usepackage{xcolor}         \usepackage{bbm}
\usepackage{pifont}
\usepackage{subcaption}

\acmYear{2024}\copyrightyear{2024}
\setcopyright{rightsretained}
\acmConference[FAccT '24]{The 2024 ACM Conference on Fairness, Accountability, and Transparency}{June 3--6, 2024}{Rio de
Janeiro, Brazil}
\acmBooktitle{The 2024 ACM Conference on Fairness, Accountability, and Transparency (FAccT '24), June 3--6, 2024, Rio de
Janeiro, Brazil}\acmDOI{10.1145/3630106.3658927}
\acmISBN{979-8-4007-0450-5/24/06}

\begin{document}

\def\ie{\textit{i.e.}, \onedot}
\def\eg{\textit{e.g.}, \onedot} 
\def\wrt{\textit{w.r.t.} \onedot}

\newcommand{\fix}{\marginpar{FIX}}
\newcommand{\new}{\marginpar{NEW}}
\definecolor{darkgreen}{rgb}{0.0, 0.5, 0.0}
\definecolor{royalblue}{RGB}{50, 92, 168}
\definecolor{coral}{rgb}{1.0, 0.5, 0.31}
\definecolor{purple}{rgb}{0.5, 0., 0.5}
\newcommand{\melissahall}[1]{\textcolor{darkgreen}{\textbf{MH:~}#1}}
\newcommand{\costest}[1]{\textcolor{coral}{\textbf{[Cost Estimate]~}#1}}
\newcommand{\candace}[1]{\textcolor{royalblue}{\textbf{CR:~}#1}}
\newcommand{\adriana}[1]{\textcolor{purple}{\textbf{ARS:~}#1}}
\newcommand{\TODO}{\textcolor{red}{\textbf{TODO~}}}
\newcommand{\notab}{\hspace{-1em}}
\newcommand{\rulesep}{\unskip\ \vrule\ }
\newcommand{\obj}{\texttt{\{object\}}\xspace}
\newcommand{\objreg}{\texttt{\{object\} in \{region\}}\xspace}
\newcommand{\objcountry}{\texttt{\{object\} in \{country\}}\xspace}
\newcommand{\syncountry}{\texttt{\{synonym\} in \{country\}}\xspace}
\newcommand{\ldmacc}{LDM 2.1\xspace}
\newcommand{\dmacc}{DM w/ CLIP\xspace}
\newcommand{\glide}{GLIDE\xspace}
\newcommand{\georep}{geographic representation\xspace}
\renewcommand{\floatpagefraction}{.8}
\newcommand{\cmark}{\ding{51}}\newcommand{\xmark}{\ding{55}}
\title{Towards Geographic Inclusion in the Evaluation of Text-to-Image Models}

\author{Melissa Hall}
\affiliation{ \institution{Meta AI}
 \country{United States}
}
\author{Samuel J. Bell}
\affiliation{ \institution{Meta AI}
 \country{France}
}
\author{Candace Ross}
\affiliation{ \institution{Meta AI}
 \country{United States}
}
\author{Adina Williams}
\affiliation{ \institution{Meta AI}
 \country{United States}
}
\author{Michal Drozdzal}
\authornote{Equal contribution in research and engineering leadership.}
\affiliation{ \institution{Meta AI}
 \country{Canada}
}
\author{Adriana Romero Soriano}
\authornotemark[1]
\affiliation{ \institution{Meta AI, Mila, McGill University, Canada CIFAR AI Chair}
 \country{Canada}
}

\renewcommand{\shortauthors}{Hall et al.}

\begin{abstract}
Rapid progress in text-to-image generative models coupled with their deployment for visual content creation has magnified the importance of thoroughly evaluating their performance and identifying potential biases.
In pursuit of models that generate images that are realistic, diverse, visually appealing, and consistent with the given prompt, researchers and practitioners often turn to automated metrics to facilitate scalable and cost-effective performance profiling.
However, commonly-used metrics often fail to account for the full diversity of human preference; often even in-depth human evaluations face challenges with subjectivity, especially as interpretations of evaluation criteria vary across regions and cultures.
In this work, we conduct a large, cross-cultural study to study how much annotators in Africa, Europe, and Southeast Asia vary in their perception of geographic representation, visual appeal, and consistency in real and generated images from state-of-the art public APIs. 
We collect over $65,000$ image annotations and $20$ survey responses.
We contrast human annotations with common automated metrics, finding that human preferences vary notably across geographic location and that current metrics do not fully account for this diversity.
For example, annotators in different locations often disagree on whether exaggerated, stereotypical depictions of a region are considered geographically representative. 
In addition, the utility of automatic evaluations is dependent on assumptions about their set-up, such as the alignment of feature extractors with human perception of object similarity or the definition of ``appeal'' captured in reference datasets used to ground evaluations. 
We recommend steps for improved automatic and human evaluations.
This includes collecting annotations from people located inside and outside the region of interest, instructing annotators on whether they should follow specific definitions of evaluation criteria or utilize their own interpretation, and reporting assumptions underlying automatic evaluations.

\end{abstract}

\ccsdesc[500]{Social and professional topics}
\ccsdesc[500]{Applied computing}
\ccsdesc[500]{Computing methodologies~Computer vision}

\keywords{text-to-image generation, geography, evaluation}

\maketitle

\begin{figure}[t]
     \centering
     \begin{subfigure}[b]{0.495\textwidth}
         \centering
         \includegraphics[width=\textwidth]{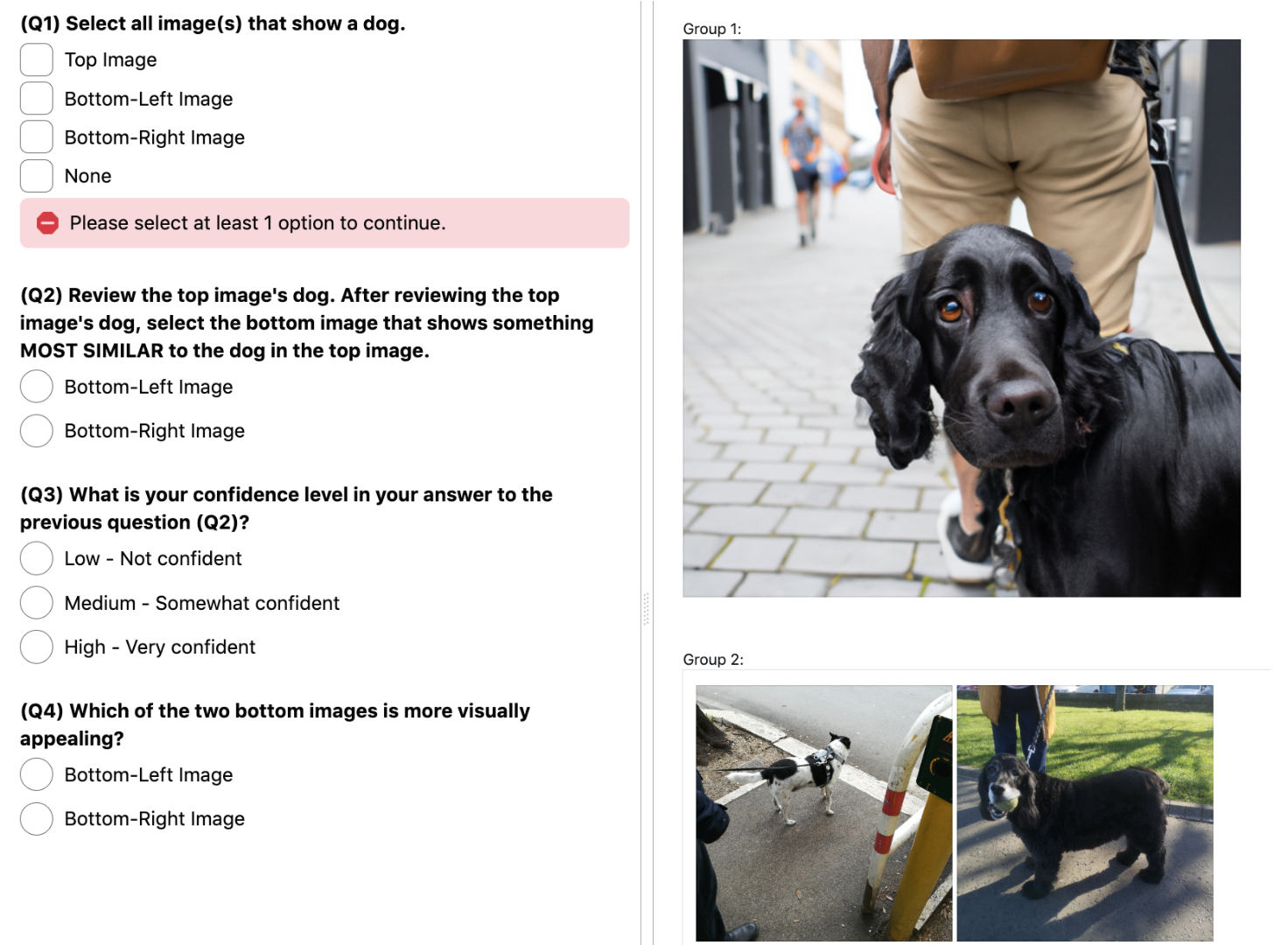}
         \caption{Task 1}
         \label{fig:t1}
     \end{subfigure}
     \begin{subfigure}[b]{0.495\textwidth}
         \centering
         \includegraphics[width=\textwidth]{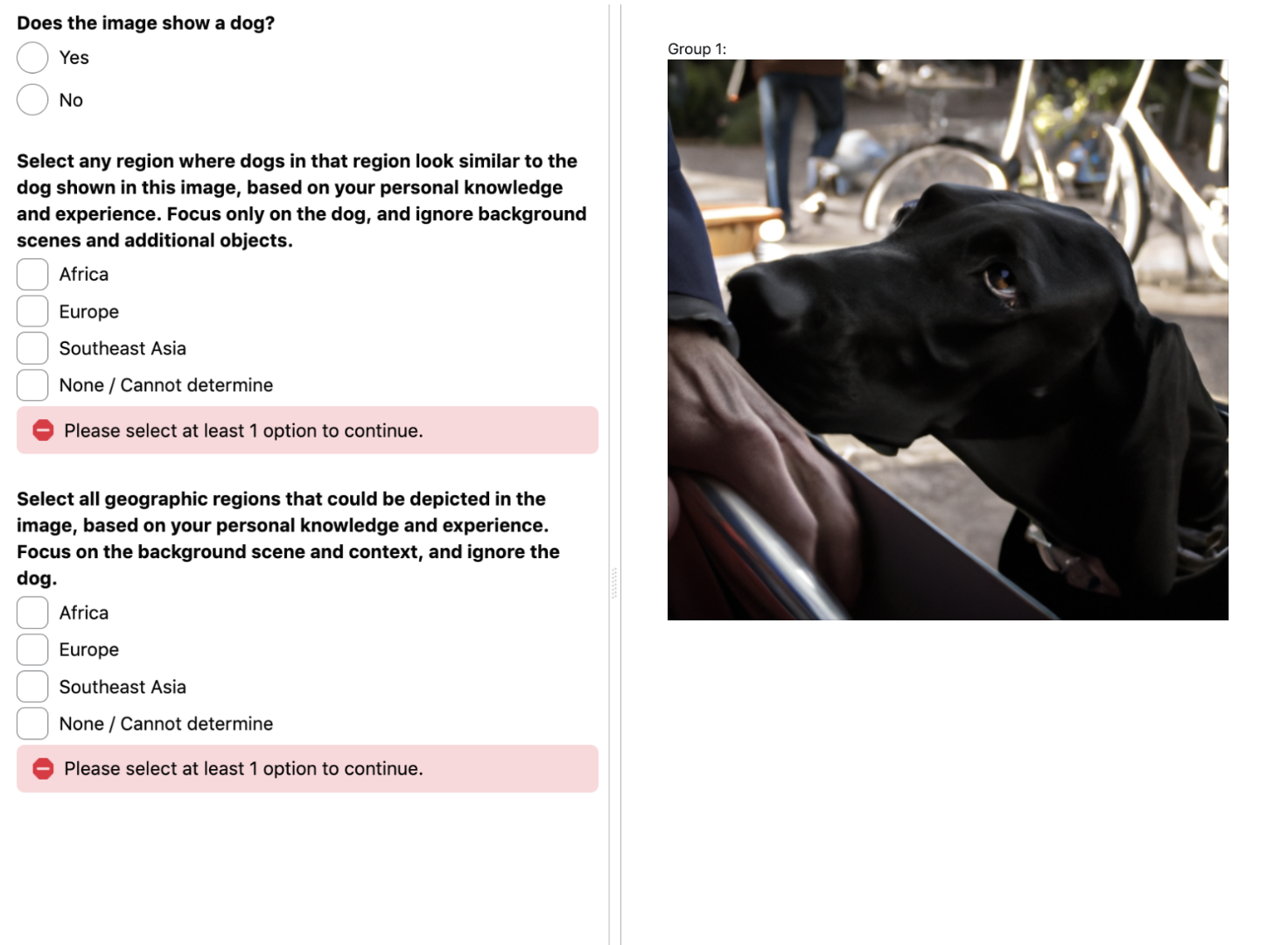}
         \caption{Task 2}
         \label{fig:t2}
     \end{subfigure}
     \caption{We collect 65k annotations performed by people located in Africa, Europe, and Southeast Asia corresponding to evaluation criteria of text-to-image models including \textbf{\georep}, \textbf{similarity}, \textbf{visual appeal}, and \textbf{object consistency} in real and generated images.      We develop recommendations for improved human and automatic evaluations of text-to-image models.}
     \label{fig:task_ex}
\end{figure}

\section{Introduction}
\label{sec:introduction}

Recent years brought unprecedented progress in generative models for visual content creation, with works achieving impressively photorealistic image generations~\citep{ramesh2022hierarchical,DBLP:journals/corr/abs-2112-10752,DBLP:journals/corr/abs-2112-10741,gafni2022make}. 
However, it is important to understand whether the true diversity of the real world is conveyed in generated images. 
In particular, generated images should be \emph{representative} of the world, capturing the myriad variability of people, objects, and scenes across geographic regions, while also \emph{visually appealing} or interesting, and \emph{consistent} with the input text description.
Assessing these properties is crucial for an in-depth understanding of the performance and potential biases of text-to-image systems. 

Commonly used automated metrics, such as the Fr\'{e}chet Inception Distance (FID)~\citep{DBLP:journals/corr/HeuselRUNKH17}, Inception Score~\citep{salimans2016improved} and precision and recall (PR)-based metrics~\citep{Sajjadi2018_PR, kynkäänniemi2019improved,DBLP:journals/corr/abs-2002-09797,Shmelkob2018,DBLP:journals/corr/abs-1905-10887}, already exist for evaluating the representativeness of generated images. Other metrics such as LPIPS~\citep{zhang2018unreasonable} and Vendi Score~\citep{friedman2022vendi} target diversity within a corpus of generated images explicitly. 
Metrics such as CLIPScore~\citep{hessel2021clipscore} have also been introduced to automatically evaluate the consistency of generated images with respect to the input text description. 
Recently, several metrics have been extended to measure the ability of text-to-image systems to create faithful depictions of objects around the globe~\citep{hall2023dig}. 

However, these metrics have notable challenges and are undergoing debates about their robustness~\citep{morozov2021on,stein2023exposing}. 
For example, the aforementioned metrics rely on pre-trained feature extractors, such as Inceptionv3~\citep{DBLP:journals/corr/SzegedyVISW15}, which can lead to unreliable model rankings and susceptibility to undesirable ``fringe features''~\citep{kynkäänniemi2023role}.
FID and PR-based metrics rely on reference datasets to define representativeness that may not capture the true diversity of the real world. 
These metrics operationalize representations in a constrained way and fail to account for the diversity of human preferences \cite{10.1145/3593013.3594016}.

Given the complexity and contestedness of automatic metrics, human evaluations remain the gold standard to benchmark text-to-image systems. 
Human evaluations often focus on side-by-side comparisons of images taken from different sources~\citep{dai2023emu,ramesh2022hierarchical,DBLP:journals/corr/abs-2112-10752,DBLP:journals/corr/abs-2112-10741} or task human auditors with identifying concepts or group information in an image based on instructional criteria ~\citep{basu2023inspecting}. 
However, human evaluations also face challenges. 
Most notably, they struggle to account for the subjectivity across regions and cultures. 
Recent studies suggest that human evaluations tend to be somewhat ad hoc, with task design impacting results and leading to high variance in estimates~\citep{DBLP:journals/corr/abs-1904-01121}. 
Furthermore, annotations of model performance and bias presuppose definitions that may not encapsulate the full range of human perceptions, may be mutually incompatible between different groups of people, and may not map to what the model creator is genuinely interested in.~\looseness-1

In this paper, we present a human study to understand how annotators conceptualize desirable properties of text-to-image systems across geographic locations. 
We focus on images of \textit{objects}, as notable  disparities have been observed between geographic regions in both text-to-image generative systems ~\citep{hall2023dig,basu2023inspecting,bianchi2022easily} and discriminative models ~\citep{devries2019does,Singh_2022_CVPR,gustafson2023pinpointing,richards2023progress}.

We collect over $65,000$ annotations with $20$ qualitative survey responses to study how people in Africa, Europe, and Southeast Asia vary in their perception of geographic representation, visual appeal, and consistency.
The study includes both real, geographically diverse images and images created by two state-of-the-art generative systems that depict six types of cross-culturally common objects. 
We contrast the results of the human annotations with commonly used automated metrics. 
Through our study, we find that: 
\begin{itemize}
    \item Perceptions of \georep, appeal, and consistency vary across annotator locations.
    \item Annotators can reinforce geographic stereotypes and contradict each other in their interpretation of appeal. 
    Preferred qualities include realism, foregrounding, exaggerated backgrounds, bold colors, and luxurious objects.
    \item Annotators vary in how much they follow concrete definitions or apply their subjective interpretation when completing tasks. Some rely on personal knowledge while others utilize external reference sources. 
    \item CLIP and DINOv2 feature extractors correlate better with human judgment of similarity than Inceptionv3.    \item Although CLIPScore is highly reliable in assessing the presence of an object in an image when used with a threshold, it is generally unreliable for assessing \georep.
\end{itemize}

In addition, we recommend steps to improve evaluations of text-to-image generative models.
To strengthen human evaluations, we suggest clearly instructing annotators about whether they should attempt to capture their subjective perspective or to follow specified definitions, and whether they should leverage external resources or rely only on their existing knowledge. 
In addition, we suggest collecting perspectives from both in-region and out-of-region annotators to capture more breadth. 
To strengthen automatic evaluations, we suggest selecting reference data sources with thoughtful consideration of the interpreted criteria and using more modern feature extractors that better capture image qualities relevant for the evaluation, such as shape.
When reporting evaluations, avoid ``majority-vote'' types of aggregations that reduce uncertainty to binary indicators of evaluation criteria, and communicate clearly the assumptions of criteria definitions used in the measurement process. 

We hope that this work strengthens evaluations of geographic representation in text-to-image generative models and paves the way towards text-to-image systems that truly work for everyone.\looseness-1

\begin{table}[]
\footnotesize
\begin{tabular}{c|c|c|c}
\textbf{Evaluation Criteria} &
  \textbf{Definition} &
  \textbf{Automatic Metric} &
  \textbf{Our Annotator Data} \\ 
  \hline
  \begin{tabular}[c]{@{}c@{}}Realism / Diversity \end{tabular} &
  \begin{tabular}[c]{@{}c@{}}Whether, for a geographic region,\\ generated samples resemble the\\  real world / capture the variability \\ of the real world\end{tabular} &
  \begin{tabular}[c]{@{}c@{}}Precision / Coverage\\ \\ Region-CLIPScore\end{tabular} &
  \begin{tabular}[c]{@{}c@{}}Geographic representation of objects: \\ "Which regions have similar objects?"\\ \\ Geographic representation of backgrounds: \\ "Which regions have similar backgrounds?"\end{tabular} \\  \cdashline{2-4}
Similarity &
  \begin{tabular}[c]{@{}c@{}}Whether two images closely \\ resemble each other\end{tabular} &
  Distance &
  \begin{tabular}[c]{@{}c@{}}Similarity:\\ "Which image has an object more\\similar to the reference image?"\end{tabular} \\ \hline
Visual appeal &
  \begin{tabular}[c]{@{}c@{}}Whether images have  visual \\ attractiveness or interest\end{tabular} &
  Precision &
  \begin{tabular}[c]{@{}c@{}}Visual appeal: \\ "Which image is more appealing?"\end{tabular} \\ \hline
Object consistency &
  \begin{tabular}[c]{@{}c@{}}Whether images show elements\\of prompt used in generation\end{tabular} &
  Object-CLIPScore &
  \begin{tabular}[c]{@{}c@{}}Object consistency: \\ "Is the object shown?"\end{tabular}
\end{tabular}
\caption{We study \textbf{\georep} (encapsulating realism/diversity and similarity), \textbf{visual appeal}, and \textbf{object consistency}. The table summarizes their definitions, metrics used in their approximation, and relevant annotation data that we collected.
}
\label{table:eval}
\end{table}

\section{Experimental Set-up}
\label{sec:set_up}

We outline evaluation criteria for text-to-image models and describe our annotations, summarized in Table \ref{table:eval}. 

\subsection{Evaluation Criteria for Text-to-Image Generative Models}
We discuss evaluation criteria for text-to-image models and automatic metrics used for their approximation. 
In later analyses, we encapsulate the realism / diversity and similarity criteria into an overall focus of \georep. 
We also study image visual appeal and object consistency.

\subsubsection{Realism / Diversity}

Realism is the degree to which generated samples \textit{resemble} the real world, and diversity is the degree to which samples capture the \textit{full variability} of the real world~\citep{Sajjadi2018_PR,kynkäänniemi2019improved,hall2023dig}.
Prior works~\citep{basu2023inspecting,hall2023dig} have utilized realism and diversity in the assessment of the extent to which the real-world variability and nuance of a given region is depicted in a corpus of generated images.
We call this ``\georep.''

\vspace{1mm}
\noindent \textit{Automatic metrics.}
Image realism and diversity are often measured with precision and coverage, respectively~\citep{Sajjadi2018_PR,kynkäänniemi2019improved,hall2023dig}. 
Precision highlights whether generated images look similar to real images, while coverage quantifies whether generated images capture the breadth of variation in real images.
Specifically, precision measures the proportion of generated images that lie within the manifold of real, reference images. 
Coverage measures the proportion of real images that have generated images nearby.
In Section~\ref{sec:results_new}, we analyze the vulnerability of these metrics to predefined feature extractors and reference manifolds through the context of similarity, which we introduce next.
In addition, the distances \textit{i.e.}, cosine similarity, between the CLIP text embedding corresponding to group terms and embedding of a generated image has been used to identify demographic groups~\citep{Cho2022DallEval} and  \georep~\citep{basu2023inspecting}.
We use the CLIPScore~\citep{hessel2021clipscore} with region information (``Region-CLIPScore'') as an indicator of \georep, analogously to ~\citet{lee2023holistic}.

\subsubsection{Similarity}
While similarity, \textit{i.e.}, whether two images closely resemble each other, is not commonly evaluated in text-to-image generative models, it is the foundation of criteria like realism and diversity. 
For an object in an image to have ``realism,'' it needs to resemble instances of the object in the real world. 
For a dataset of images to display enough ``diversity,'' the objects in the images must meaningfully differ from each other.

\vspace{1mm}
\noindent \textit{Automatic metric.}
Measurements of similarity are closely related to automatic evaluations of realism and diversity, as they approximate relative distances between images in a predefined feature space. 
In Section~\ref{sec:results_new}, we study how the older, more ubiquitous Inceptionv3~\citep{DBLP:journals/corr/SzegedyVISW15} compares to the more recent CLIP ViT-B/32~\citep{radford2021learning} and DINO ViT-L/14~\citep{oquab2023dinov2} feature extractors trained on larger data sources in approximating human perceptions of similarity. 

\subsubsection{Visual Appeal}
Visual appeal corresponds to whether images have visual attractiveness or interest and has become an increasingly common evaluation criteria~\citep{saharia2022photorealistic,dai2023emu,podell2023sdxl}. 
We study how automatic metrics capture subjectivity in appeal across geographic regions.

\vspace{1mm}
\noindent \textit{Automatic metric.}
According to some previous works, appeal can be automatically measured by comparing to a manifold of real images with the precision metric~\citep{kynkäänniemi2019improved,hall2023dig}.
In Section~\ref{sec:results_new}, we study whether generated images that fall in the real image manifold align with human perceptions of visual appeal.

\subsubsection{Object Consistency}
Consistency corresponds to whether an image includes all components of its prompt. 
This relates to visual concreteness \citep{brysbaert2014concreteness} and the relationship between a concept's meaning and its human-perceptible form \citep{border}. 
Per \citep{hall2023dig}, we focus on objects included in the prompt, \textit{e.g.}, whether an image generated with ``car in Africa'' depicts a car.~\looseness-1

\vspace{1mm}
\noindent \textit{Automatic metric.}
In Section~\ref{sec:results_new} we explore the CLIPScore~\citep{hessel2021clipscore}, which can be used to measure the input-output consistency of the generative model \citep{lee2023holistic,saharia2022photorealistic,hall2023dig}.
Following \cite{hall2023dig}, we use the text embedding corresponding to the object that should be present in the image and refer to the metric as ``Object-CLIPScore.''

\begin{figure}[!t]
    \centering
     \begin{subfigure}[b]{0.49\textwidth}
         \centering
         \includegraphics[width=\textwidth]{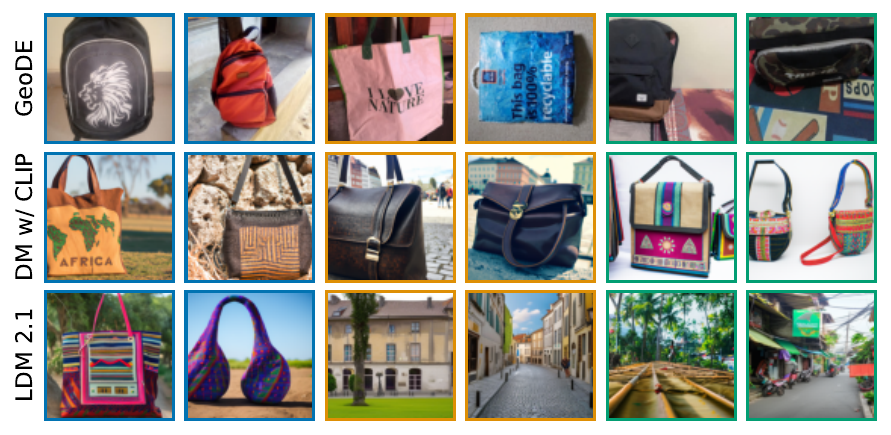}
         \caption{Bag}
     \end{subfigure}
     \hfill
     \begin{subfigure}[b]{0.49\textwidth}
         \centering
         \includegraphics[width=\textwidth]{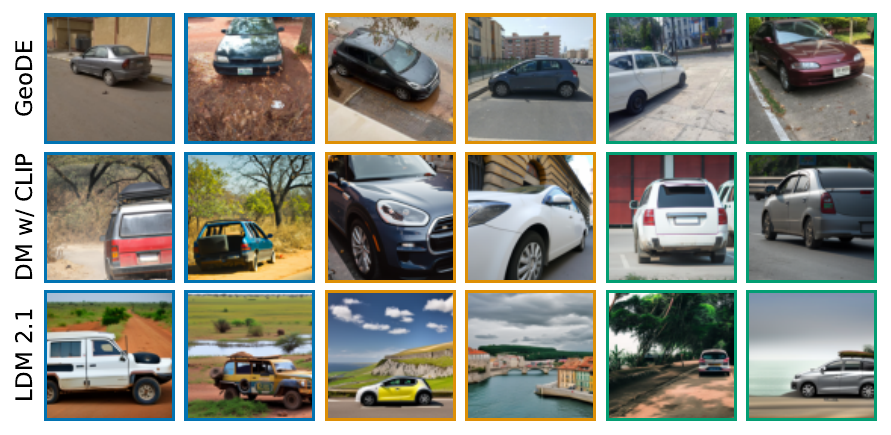}
         \caption{Car}
     \end{subfigure}
     \begin{subfigure}[b]{0.49\textwidth}
         \centering
         \includegraphics[width=\textwidth]{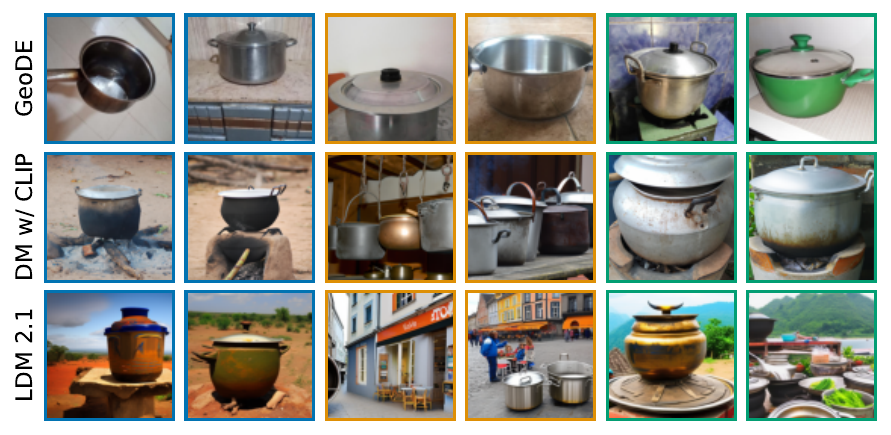}
         \caption{Cooking pot}
     \end{subfigure}
     \hfill
     \begin{subfigure}[b]{0.49\textwidth}
         \centering
         \includegraphics[width=\textwidth]{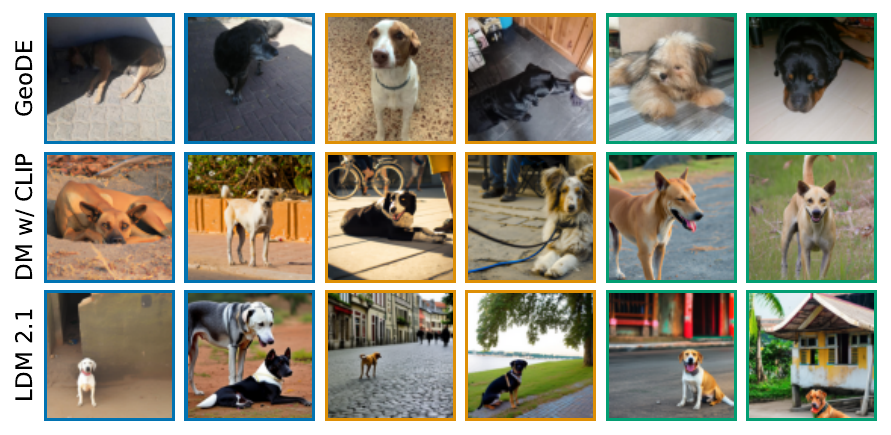}
         \caption{Dog}
     \end{subfigure}
     \begin{subfigure}[b]{0.49\textwidth}
         \centering
         \includegraphics[width=\textwidth]{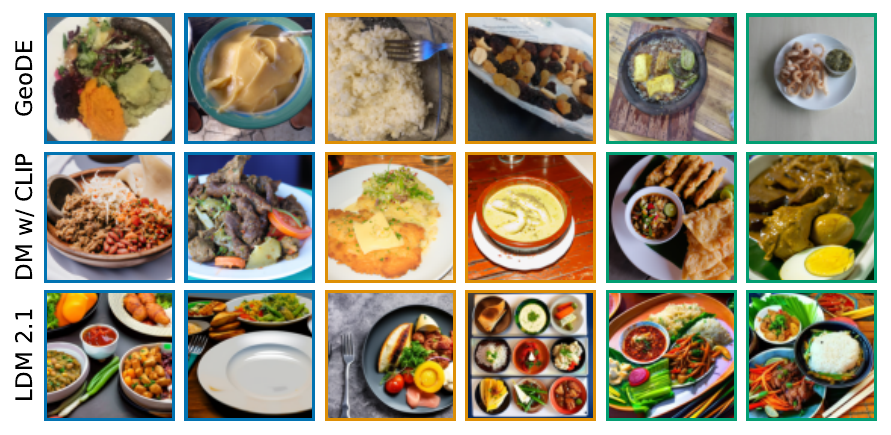}
         \caption{Plate of food}
     \end{subfigure}
     \hfill
     \begin{subfigure}[b]{0.49\textwidth}
         \centering
         \includegraphics[width=\textwidth]{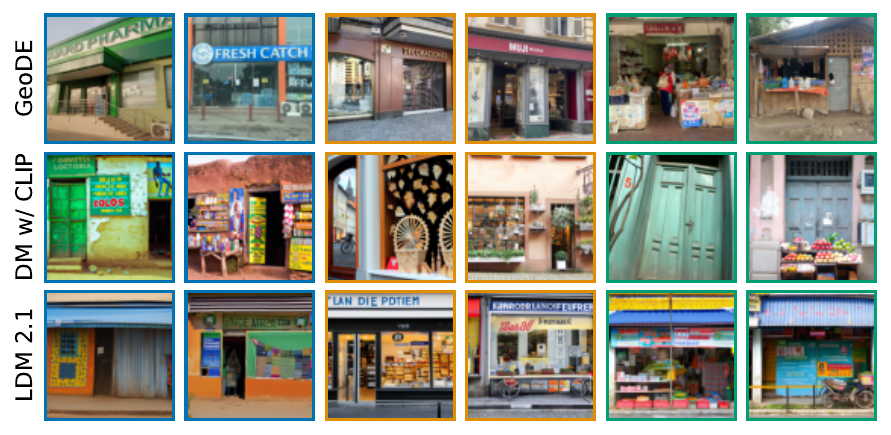}
         \caption{Storefront}
     \end{subfigure}
     \vspace{-3mm}
    \caption{Random examples of real images from GeoDE in each region and generated images from \dmacc and \ldmacc using the prompt \objreg. The first two columns correspond to \textcolor{blue}{Africa}, the next to \textcolor{orange}{Europe}, and the last to  \textcolor{darkgreen}{Southeast Asia}.}
    \label{fig:example_mosaic}
\end{figure}

\subsection{Annotations}

We now describe the annotations we collected to better understand how humans in different geographic locations vary in their perception of the aforementioned evaluation criteria for text-to-image models.

\subsubsection{Images}

Our tasks include real images from GeoDE~\citep{ramaswamy2022geode}, a diverse, geographically representative dataset of images taken across multiple regions. 
Each image contains an object filling at least 25\% of the image and occurs in everyday settings.
We also include generated images from a multi-modal implementation of a generative pre-trained transformer leveraging CLIP image embeddings. This model, which we call ``\dmacc'', has approximately 3.5 billion parameters~\citep{ramesh2022hierarchical}.
We also use a latent diffusion model trained on a public dataset of approximately 5 billion images, excluding explicit material~\citep{DBLP:journals/corr/abs-2112-10752}, which we refer to as ``\ldmacc.''
We use paid versions of both models and generate images using the prompts ``\obj'' and ``\objreg'' to  capture representations across regions.

We focus on six objects (\textit{bag}, \textit{car}, \textit{cooking pot}, \textit{dog}, \textit{plate of food}, and \textit{storefront}) that exist across regions \cite{ramaswamy2022geode} and have been used to audit generative and discriminative models~\citep{hall2023dig,devries2019does,richards2023progress,bansal2022how}.
We include the regions \textit{Africa}, \textit{Europe}, and \textit{Southeast Asia}. 
Generated images of these objects vary in their geographic representation and consistency ~\citep{hall2023dig}, as shown in randomly selected images in Figure~\ref{fig:example_mosaic}. 
For example, generated images of \textit{car}, \textit{cooking pot} and \textit{storefront} often depict geographic stereotypes not seen in the real dataset, such as rudimentary infrastructure, low-income tools, and vibrant colors for Africa and Southeast Asia. 
Similarly, generated images have exaggerated backgrounds, with rural and cobblestone/alley-filled scenes for Africa and Europe, respectively, that are rarely depicted in GeoDE.
Furthermore, \ldmacc shows consistency issues with intended objects like \textit{bag}, \textit{car}, and \textit{cooking pot} often left out of generated images.

\subsubsection{Tasks}
Our study comprises two main annotation tasks that we populate with different image combinations, balanced across objects and regions. Figure~\ref{fig:task_ex} shows examples of the user interface for Task 1 and Task 2.

\vspace{1mm}
\noindent \textit{Task 1: Image Comparison \& Object Consistency.}
The first task, in Figure \ref{fig:t1}, focuses on understanding object consistency and human perceptions of image similarity and appeal.
As with classical ABX testing \citep{munson1950standardizing}, we display triplets of a single reference image and two images for comparison, one real and one generated. 
The annotator is asked to indicate whether the intended object is shown in each image, providing insight into their perception of \textit{object consistency}.
Then, they are asked to select which of the two comparison images shows an object more similar to the object in the reference image and to provide a confidence score to their answer. 
This allows us to measure potential variations in annotator perception of \textit{similarity}.
Finally, annotators decide which comparison image is more \textit{visually appealing}.

For Task 1, images in a given triplet have the region in common or are generated with the \texttt{object} prompt.
Generated images in a triplet are from the same model.
Across triplets, the comparison images vary in their distance from each other and from the reference images in the Inceptionv3 feature space. 
We use Inceptionv3 as it is well-established when assessing performance of generative models~\citep{heusel2018gans,casanova2021instance,ramesh2022hierarchical,DBLP:journals/corr/abs-2112-10752} (and we study additional feature extractors in our analysis of similarity).
We define the triplet distance as $||f(I_{ref})-f(I_{far})||^2 -||f(I_{ref})-f(I_{close})||^2$, where $f$ represents the feature extractor, $I_{ref}$ is the reference image, $I_{close}$ is the closer comparison image to the reference in the embedding space, and $I_{far}$ is the further image. 
A higher triplet distance means that one comparison image is much further from the reference image than the other comparison image, suggesting it is much more ``similar'' to the reference image.
For every model, we compile 425 triplets for each object-region combination, yielding 15,300 triplets. 
Appendix~\ref{app:set_up} contains additional details about the triplets.

\vspace{1mm}
\noindent \textit{Task 2: Geographic Representation \& Object Consistency.}
The second task, in Figure \ref{fig:t2}, focuses on understanding consistency and \georep of objects and scenes in images. 
A single image is shown, and annotators are asked to indicate if the intended object is present. 
If so, they are then asked to select any region from a pre-specified list that they feel the object could be from, i.e., that they feel has objects from the same category (e.g. ``bag'', ``car'') that look similar to the object in the image. 
They are also asked the same question about the background scene / context depicted in the image. 
These questions correspond to annotator perception of \georep.

For Task 2, we randomly select 100 images from GeoDE for each region-object combination. 
In addition, for \ldmacc and \dmacc, we used 100 images with generated \objreg for each object-region combination and 100 images generated with \obj for each object. 
This yields 6,600 images.

\subsubsection{Annotators}
Each task is annotated by three people from our pool: one person located in Africa, Europe, and Southeast Asia.
The annotator pool consists of 15-25 people in each region. After both tasks are completed, we filter out approximately 13\% of Task 1 annotations and 2\% of Task 2 annotations that do not meet quality checks on real images.
The remaining annotations yield the corpus with which we perform our analyses.
Appendix~\ref{app:set_up} describes the annotator training process and compensation, quality criteria, and statistics about annotator background and job execution.

\subsection{Annotator Survey}
Once all annotations were collected, we also elicited voluntary survey responses to learn more about how annotators interpreted questions related to \georep.
The first set of questions focuses on the depiction of objects.
We ask three versions of the question \textit{``When performing these tasks, how did you determine if an object looked like similar objects in [region]? If possible, please provide specific examples,''} where we consider Africa, Europe, and Southeast Asia. 
We also ask, \textit{``Did you ever mark that none of the regions had similar objects as the ones shown in the image, or that you could not determine? If so, what prompted you to make that decision?''}
For the second set of questions, we asked the same questions, but about the \textit{``background scene / context''} of the image. 
We received 23 responses to the annotator survey, from 5 annotators in Africa, 8 in Europe, and 10 in Southeast Asia. 
We applied an inductive coding approach to categorize themes discussed by the annotators, resulting in 16 descriptive codes. 
After an initial round to validate the coding scheme, responses were individually coded by three of the authors.
We binarize the presence of each code per response and evaluate inter-coder agreement using Fleiss' Kappa~\citep{fleiss1971mns}.
We discard codes with a coefficient less than 0.6 to ensure consistent deployment by coders.
Final codes were assigned via a majority vote over the three authors, and responses were grouped according to whether they correspond to background scene or object. All codes and their Fleiss' Kappa coefficients can be found in  Appendix Table~\ref{table:survey-codes}.  
\section{Results}
\label{sec:results_new}

For each criteria, we first present analyses corresponding to \textit{Human Interpretations}, leveraging only annotation and survey data (where relevant). 
Then we discuss the \textit{Human \& Automatic Metric Interaction}, where we analyze the extent to which automatic metrics capture human interpretation of evaluation criteria.
Along the way, we suggest recommendations for improved human and automatic evaluations of text-to-image models based on our findings.

\subsection{Geographic Representation}

\subsubsection{Human Interpretations: Realism / Diversity}

\begin{figure}
    \centering
    \begin{subfigure}[b]{0.32\textwidth}
                             \includegraphics[width=0.95\textwidth]{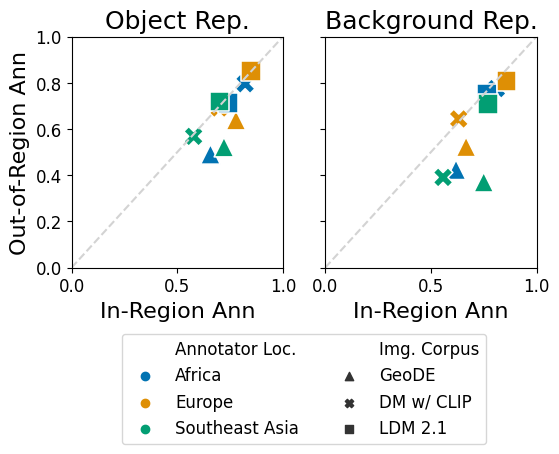}
         \vspace{8mm}
    \caption{Prop. of imgs w/ geographic rep.}
    
    \label{fig:rep_human_quant}
    \end{subfigure}
    \begin{subfigure}[b]{0.67\textwidth}
        \begin{subfigure}[b]{\textwidth}
        \includegraphics[width=0.99\textwidth]{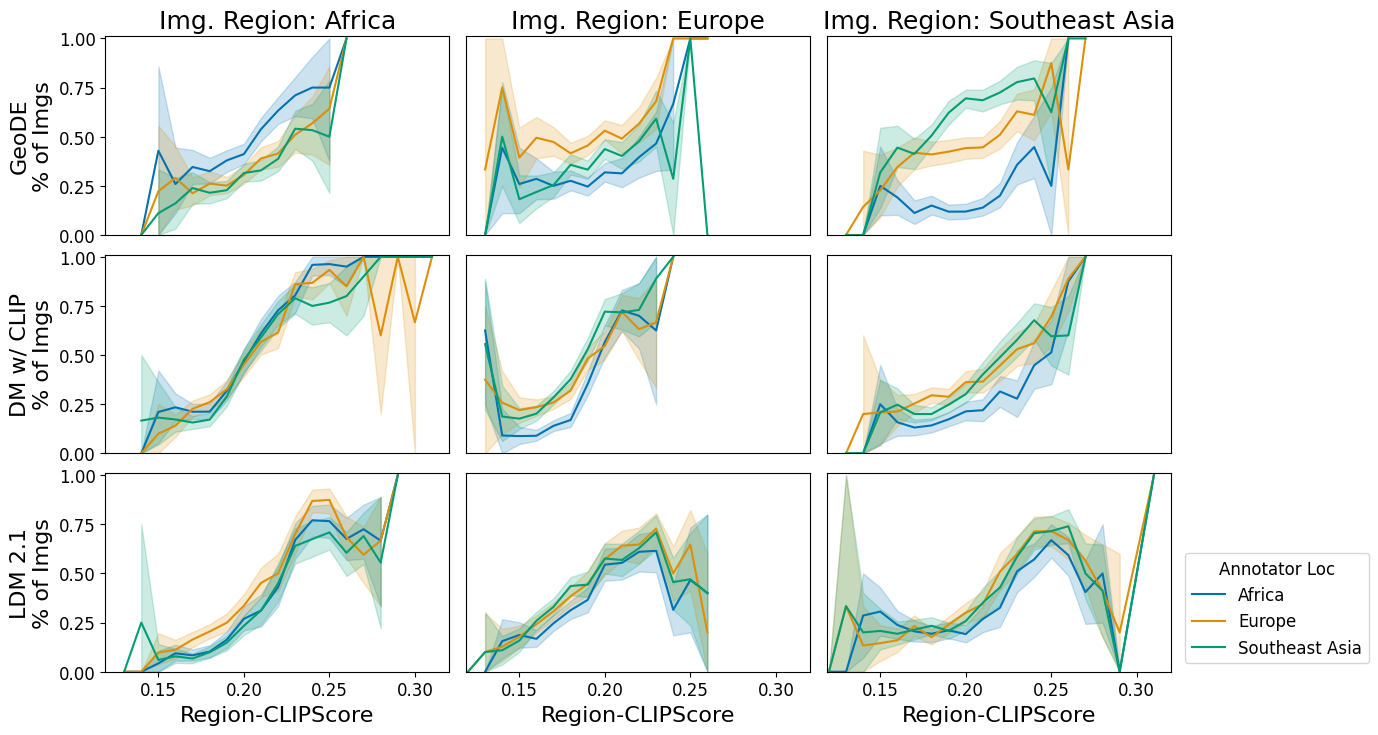}
        \end{subfigure}
        \caption{Geographic representation in background vs. Region-CLIPScore}
        \label{fig:rep_human_auto_quant_a}
    \end{subfigure}
    \caption{
    \textbf{(a)} Proportion of images that in-region and out-of-region annotations consider depicting \georep for \textit{objects} and \textit{backgrounds}. 
    \textbf{(b)} Relationship between Region-CLIPScore and annotator designation of \georep.
    The x-axis shows the average CLIPScore for all images within a bucket with size 0.01. The y-axis shows the proportion of images where annotators said the object was present. 
    We include 95\% confidence intervals generated via bootstrapping.
    \textbf{Annotator perceptions of \georep differ according to whether the annotators are located in the region of focus or outside it. 
    Region-CLIPScore does not always capture variations in perceived \georep across annotator location.}}
    \label{fig:rep_quant}
\end{figure}

\setlength{\tabcolsep}{12pt}

\begin{table}
\footnotesize
\centering
\begin{tabular}{p{3.6cm}p{6.1cm}p{2cm}}
\toprule
Theme                          & Quote                                                                                 & Re: Region    \\
\midrule
Built environment              & \textbf{R5}: ``The types of houses are really distinct''                                       & Africa \\
Culture, art \& religion       & \textbf{R6}: ``distinctive religious architecture such as Buddhist temples or mosques''        & Southeast Asia \\
Absence of detail              & \textbf{R11}: ``simply ground or tile floor with international design''                        & NA \\
External search                & \textbf{R16}: ``Sometimes I used Google picture to search the photo''                        & Africa \\
Personal lived experience      & \textbf{R5}: ``I'm familiar with European environments so I compare it to them''               & Europe \\
Nature \& natural world        & \textbf{R6}: ``Presence of tropical vegetation such as palm trees or bamboo''                  & Southeast Asia \\
People                         & \textbf{R7}: ``Black and brown people in the background''                                      & Africa \\
\midrule
Stereotyped representations    & \textbf{R19}: ``If place is classy or elegant''                                                & Europe \\
Stereotyped representations    & \textbf{R7}: ``houses were not developed and tools surrounding the objects were rudimental''   & Africa \\
Stereotyped representations    & \textbf{R6}: ``traditional houses like stilt houses''                                          & Southeast Asia \\

\bottomrule
\end{tabular}
\caption{Selected survey responses about geographic representation for themes that satisfy our inter-annotator agreement threshold. ``R[N]'' indicates respondent ID. ``Re: Region'' indicates the region specified in the question. \textbf{Annotators deploy positive and negative stereotypes and use personal experience and external resources when considering geographic information in images.}}
\label{table:human_survey}
\vspace{-5mm}
\end{table}

We first analyze how perceptions of \georep in objects and backgrounds depicted in images differ based on where the annotator is located. 
In Figure~\ref{fig:rep_human_quant}, we focus on real images that were taken in a specific region and generated images created with prompts mentioning that region.
We compare perceptions of \georep between in-region and out-of-region annotators. 
We find that annotators located in the same region as a \textit{real} image tend to identify both the object and the background as representative of that region moreso than annotators located in other regions.
However, perceptions between in-region and out-of-region annotators for \textit{generated} images are more mixed.

\begin{figure}
     \centering
         \begin{subfigure}[b]{0.245\textwidth}
             \centering
             \includegraphics[width=\textwidth]{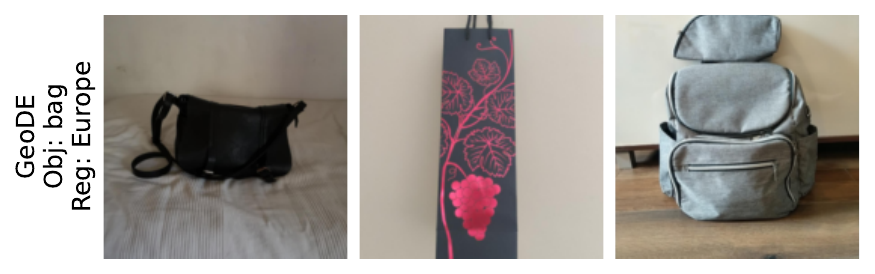}
         \end{subfigure}
         \begin{subfigure}[b]{0.245\textwidth}
             \centering
             \includegraphics[width=\textwidth]{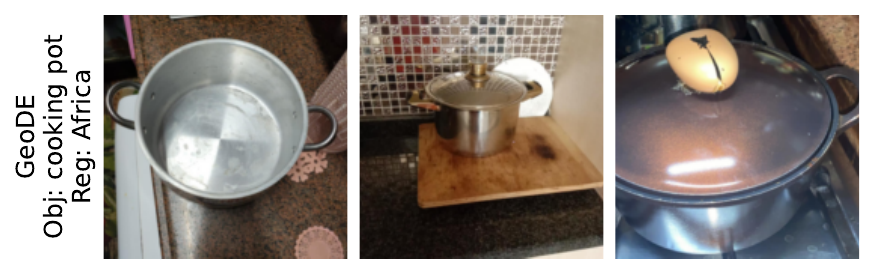}
         \end{subfigure}
         \begin{subfigure}[b]{0.245\textwidth}
             \centering
             \includegraphics[width=\textwidth]{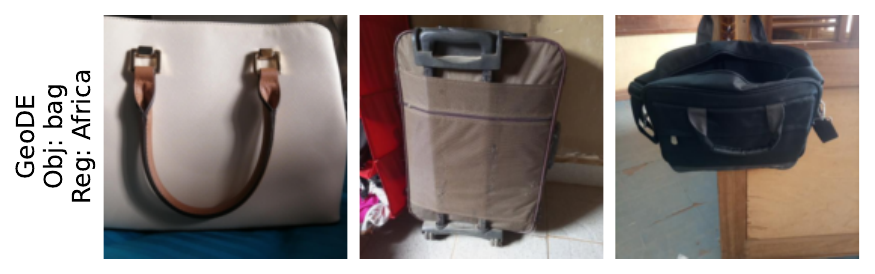}
         \end{subfigure}
         \begin{subfigure}[b]{0.245\textwidth}
             \centering
             \includegraphics[width=\textwidth]{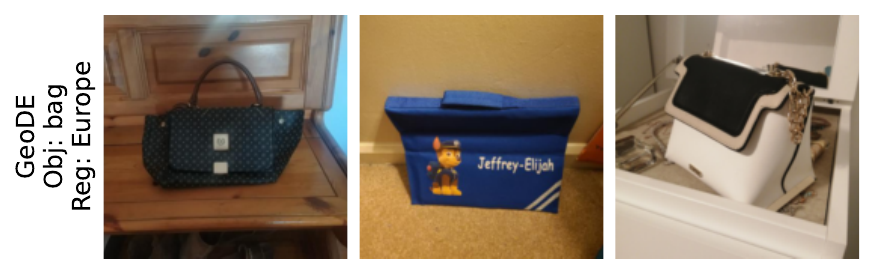}
         \end{subfigure}
         \begin{subfigure}[b]{0.245\textwidth}
             \centering
             \includegraphics[width=\textwidth]{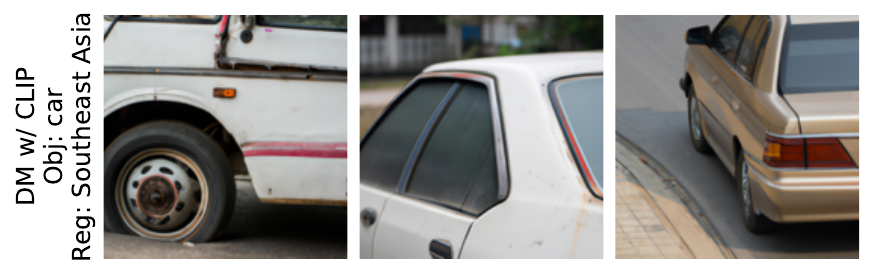}
         \end{subfigure}
         \begin{subfigure}[b]{0.245\textwidth}
             \centering
             \includegraphics[width=\textwidth]{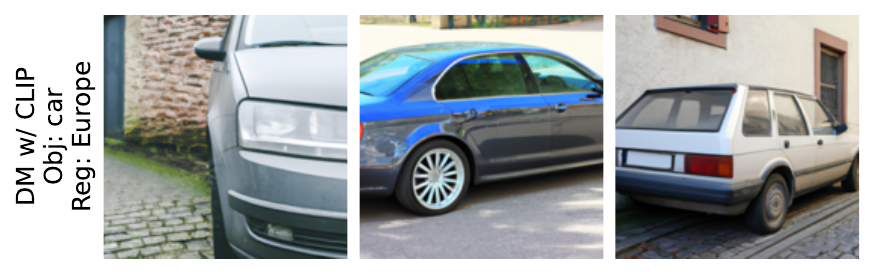}
         \end{subfigure}
         \begin{subfigure}[b]{0.245\textwidth}
             \centering
             \includegraphics[width=\textwidth]{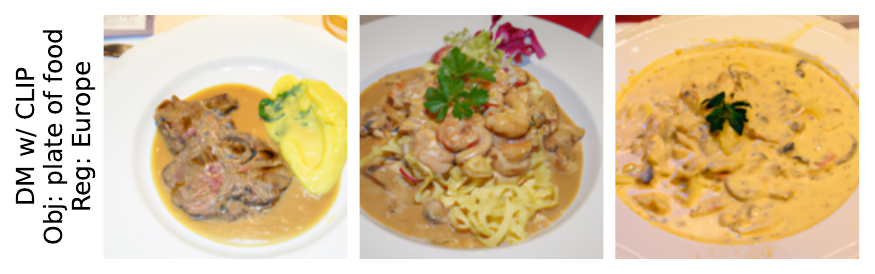}
         \end{subfigure}
         \begin{subfigure}[b]{0.245\textwidth}
             \centering
             \includegraphics[width=\textwidth]{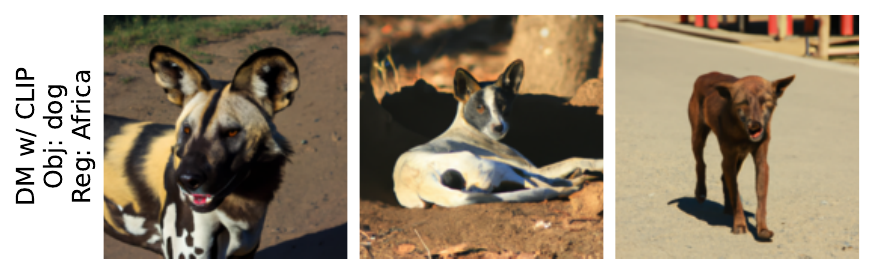}
         \end{subfigure}
         \begin{subfigure}[b]{0.245\textwidth}
             \centering
             \includegraphics[width=\textwidth]{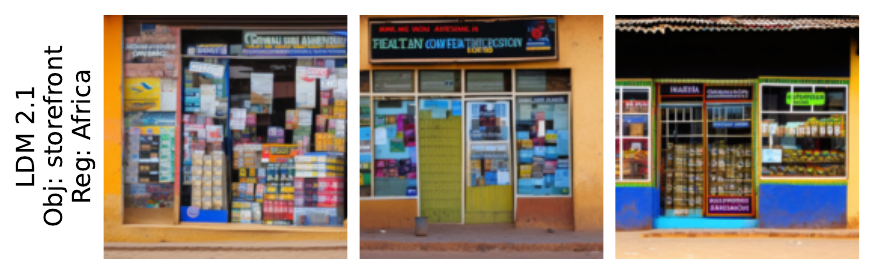}
             \caption{In: Obj is representative\newline
             Out: Obj is not representative}
             \label{fig:rep_human_visual_a}
         \end{subfigure}
         \begin{subfigure}[b]{0.245\textwidth}
             \centering
             \includegraphics[width=\textwidth]{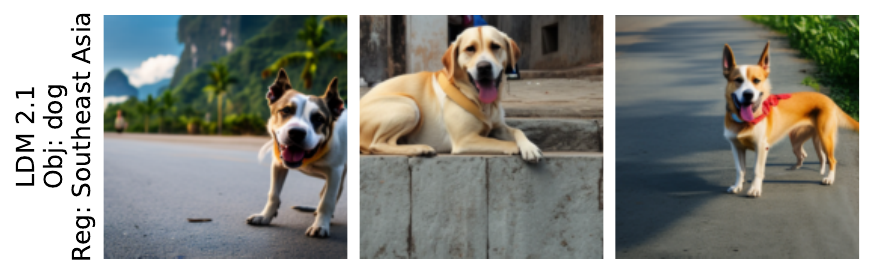}
            \caption{In: Back is representative\newline
             Out: Back is not representative}
             \label{fig:rep_human_visual_c}
         \end{subfigure}
         \begin{subfigure}[b]{0.245\textwidth}
             \centering
             \includegraphics[width=\textwidth]{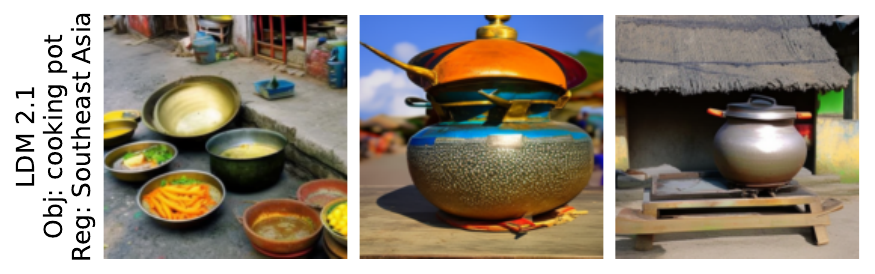}
              \caption{In: Obj is not representative\newline
             Out: Obj is representative}
             \label{fig:rep_human_visual_b}
         \end{subfigure}
         \begin{subfigure}[b]{0.245\textwidth}
             \centering
             \includegraphics[width=\textwidth]{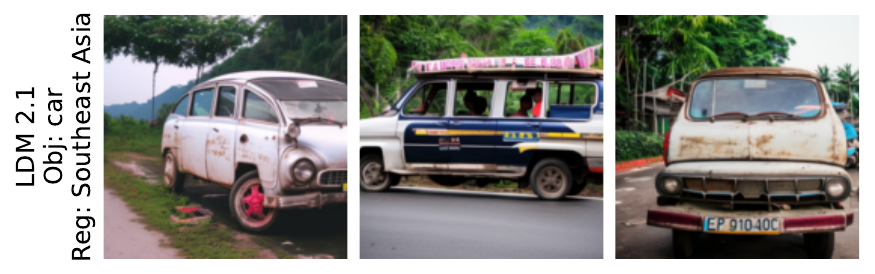}
            \caption{In: Back is not representative\newline
             Out: Back is representative}
             \label{fig:rep_human_visual_d}
         \end{subfigure}
     \caption{
     Examples of disagreement between in- and out-of-region annotators about \georep of \textit{objects} and \textit{backgrounds}.
     \textbf{Out-of-region annotators tend to consider stereotypes representative more than in-region annotators.}}
     \label{fig:rep_human_visual}
\end{figure}

To better understand why in- and out-of-region annotators disagree, we include example images in Figure~\ref{fig:rep_human_visual}.
In-region annotators often consider objects and backgrounds with few geographically-distinguishing features as representative more than out-of-region annotators.  
For example, this is the case for real images depicting bags without geographically-associated patterns (Figure~\ref{fig:rep_human_visual_a}).
It is also true of generated images of cars and dogs with backgrounds dominated by agnostic outdoor streets (Figure~\ref{fig:rep_human_visual_c}).
Meanwhile, when generated images show stereotypical objects, such as simple cooking pots for Southeast Asia from \ldmacc and saucy pastas and ``meat-and-potato'' meals for Europe from \dmacc (as in Figure~\ref{fig:rep_human_visual_b}), out-of-region annotators often consider the images more representative than in-region annotators do. 
Furthermore, stereotypically rural scenes and rudimentary infrastructure are common in images of Africa and Southeast Asia that out-of-region annotators consider geographically representative and in-region annotators do not (Figure~\ref{fig:rep_human_visual_d}). ~\looseness-1

The results of the qualitative survey (see Table~\ref{table:human_survey} for a summary) also suggest an over-reliance on stereotypical image features among out-of-region annotators, \textit{e.g.}, the expectation that European backgrounds be ``classy or elegant'' (respondent R19, located in Southeast Asia), or that African ``houses were not developed'' (R7, located in Europe). 
We find that annotators report using a range of image features in determining \georep in both background scene and object, with a greater proportion of Europe-based respondents relying on features including buildings, nature, and people than annotators based in other locations. 
In addition, when asked about the \georep of objects and background scenes, annotators frequently refer to common stereotypes when explaining their judgments, such as ``stilt houses'' for Southeast Asia.
Finally, some annotators reported using external references such as ``Google picture to search the photo,'' which can introduce additional biases from online web data.
We quantify these findings according to annotator location in the Appendix (Figure ~\ref{fig:rep_human_survey_all}).

\vspace{1mm}
\noindent\fbox{    \parbox{0.98\textwidth}{        \textbf{Recommendation}: When annotating for \georep, include in-region and out-of-region annotators: In-region annotators tend to indicate realistic, visual qualities representative of a region while out-of-region annotators often indicate unique, exaggerated identifiers of a region. Contrasting these perspectives can illuminate regional stereotyping in generated images.
    }}

\subsubsection{Human Interpretations: Similarity}
\begin{figure}
     \centering
     \begin{subfigure}[b]{0.27\textwidth}
         \centering
         \includegraphics[width=\textwidth]{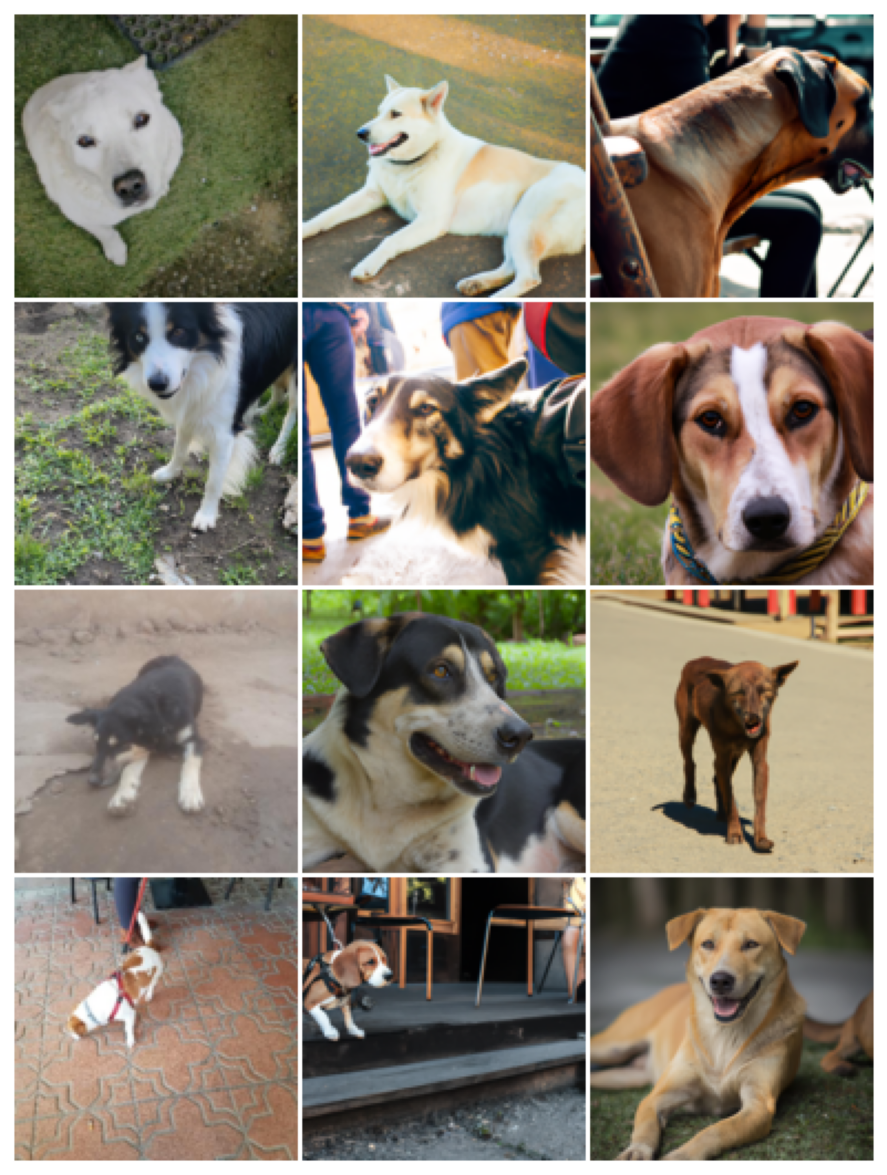}
         \caption{Consistent similarity.}
         \label{fig:qualititive_agreement_similarity}
     \end{subfigure}
          \begin{subfigure}[b]{0.27\textwidth}
         \centering
         \includegraphics[width=\textwidth]{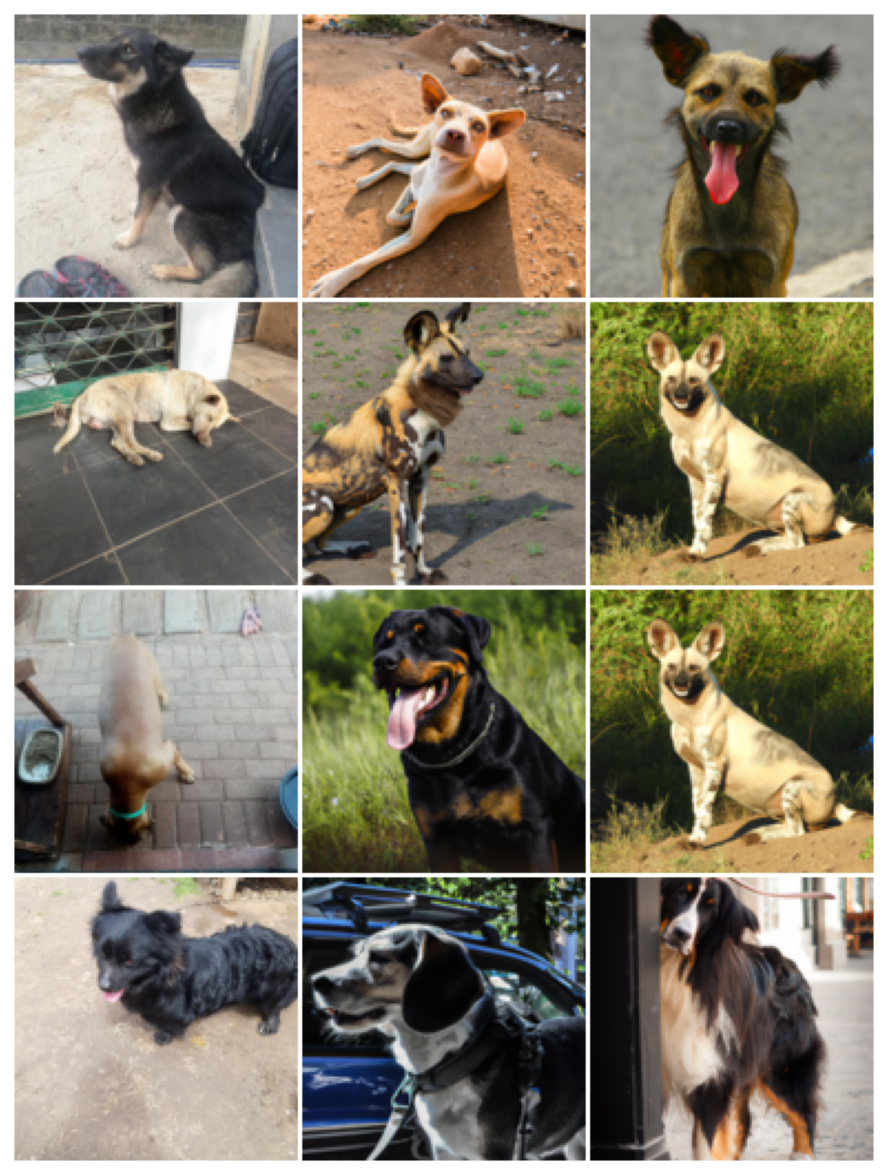}
         \caption{Inconsistent similarity}
         \label{fig:qualititive_no_agreement_similarity}
     \end{subfigure}
     \begin{subfigure}[b]{0.4\textwidth}
         \centering
         \includegraphics[width=\textwidth]{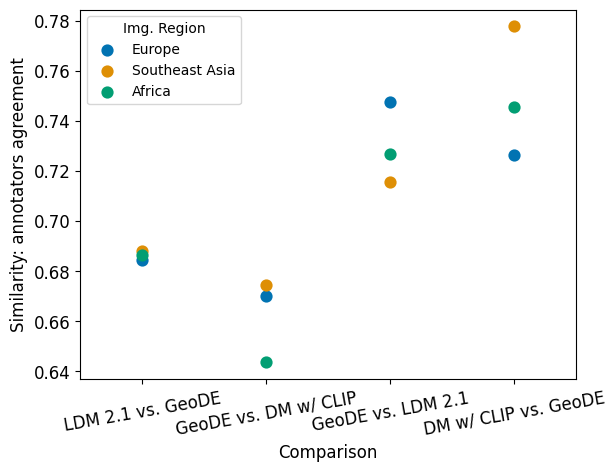}
         \caption{Similarity}
         \label{fig:agreement_similarity}
     \end{subfigure}
             \caption{
        \textbf{(a-b)} Qualitative examples of consistent and inconsistent annotator perceptions of similarity, where (a) depicts (from left): reference image from GeoDE and comparison images from \dmacc designated as more and less similar by all three annotators and (b) depicts (from left): reference image from GeoDE and two comparison images from \dmacc with inconsistent similarity annotations.
        \textbf{(c)} Rates of annotator agreement in similarity. 
        \textbf{Variations in perception of object similarity can depend on the diversity of the images. Factors in perception of similarity include color, size, type (\textit{e.g.}, dog breed), and camera angle.}}
        \label{fig:similarity}
\end{figure}

We also study how annotators vary in their assessment of image similarity.  Figures~\ref{fig:qualititive_agreement_similarity} and~\ref{fig:qualititive_no_agreement_similarity} show examples of image triplets for which the annotators agree and disagree, respectively, on which comparison image is more similar to the reference image.
We find that factors in perceptions of object similarity include color, size, type (\textit{e.g.}, dog breed), and camera angle. 

In addition, in Figure~\ref{fig:agreement_similarity}, we stratify rates of annotator agreement in similarity by the source of the triplet images and the region where the image was taken / prompted with.
We do not find that annotators agree on similarity more consistently for any image region.
However, annotators agree the least when determining which \dmacc comparison images are more similar to a GeoDE reference image.
But when this is swapped, and they are selecting between two GeoDE images compared to a single \dmacc image, they have a much higher rate of agreement. 
We hypothesize that this asymmetry may be due to the relatively low conditional diversity between the \dmacc images as compared to GeoDE (observed in Figure~\ref{fig:example_mosaic}): Selecting between less diverse images can make agreeing on the more similar image challenging, whereas two very different comparison images may allow for easier consensus about which is more similar.
Such variation in annotator perception recalls prior work suggesting alternatives to majority vote aggregations of annotations \citep{DBLP:journals/corr/abs-2110-05719}. ~\looseness-1

\vspace{1mm}
\noindent\fbox{    \parbox{0.98\textwidth}{        \textbf{Recommendation}: Collect multiple annotations for image similarity.         Aggregations like majority-vote may mask the strength of disagreement between annotators or obfuscate factors relevant to the minority perspective.
    }}

\subsubsection{Human \& Automatic Metric Interaction: Region-CLIPScore}

We next analyze how Region-CLIPScore approximates perceptions of \georep.
Higher CLIPScore values indicate stronger similarity between the image embedding and corresponding text embedding.
For each region, we calculate the Region-CLIPScore for all images and compare it to the proportion of images that annotators designate as representative. 
Results are shown in Figure ~\ref{fig:rep_human_auto_quant_a}.

We find that Region-CLIPScore tends to correspond to a larger proportion of real images perceived as representative of that region by in-region annotators than out-of-region annotators. 
Furthermore, Region-CLIPScore has a larger disparity in its approximation of \georep across annotator locations when used for identifying Southeast Asia representation than Africa and Europe.
In addition, Region-CLIPScores can be $2-5$ points higher for generated images of Africa and Southeast Asia, which tend to show strong regional stereotyping, than for Europe and real images. 
This suggests that Region-CLIPScore is susceptible to stereotypical representations of regions.

\vspace{1mm}
\noindent\fbox{    \parbox{0.98\textwidth}{        \textbf{Recommendation}: Avoid  Region-CLIPScore or use it with caution: it does not capture variation in perception of \georep between annotator locations and may favor stereotypes over realistic images.
    }}

\subsubsection{Human \& Automatic Metric Interaction: Distance}
\begin{figure}
     \centering
     \begin{subfigure}[b]{0.3\textwidth}
         \centering
         \includegraphics[width=\textwidth]{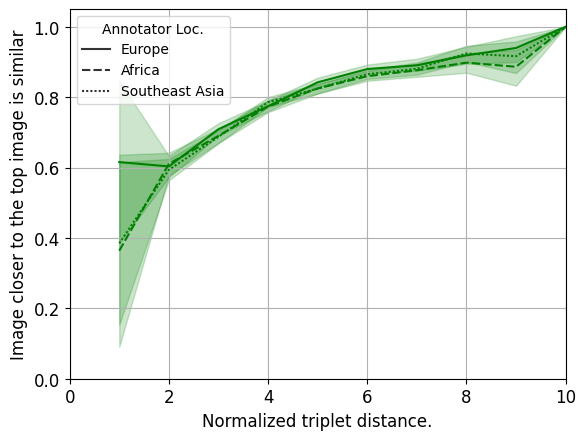}
         \caption{InceptionV3}
         \label{fig:triplet_distance_Inception}
     \end{subfigure}
     \hfill
     \begin{subfigure}[b]{0.3\textwidth}
         \centering
         \includegraphics[width=\textwidth]{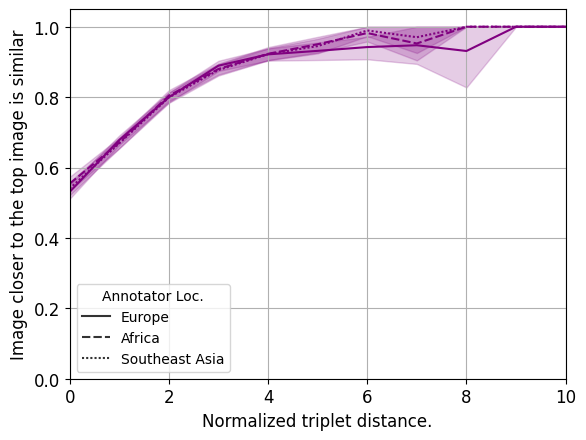}
         \caption{CLIP}
         \label{fig:triplet_distance_CLIP}
     \end{subfigure}
     \hfill
     \begin{subfigure}[b]{0.3\textwidth}
         \centering
         \includegraphics[width=\textwidth]{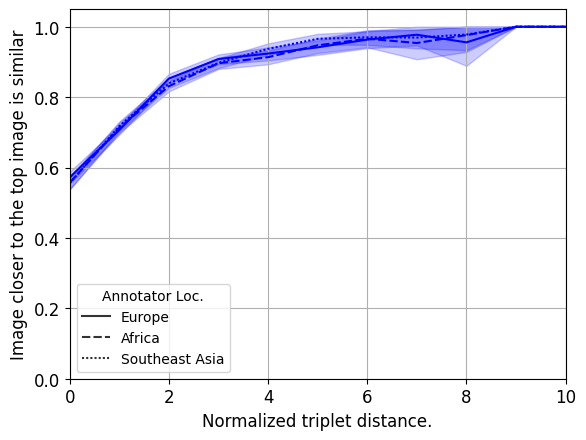}
         \caption{DINOv2}
         \label{fig:triplet_distance_DINO}
     \end{subfigure}
        \caption{Frequency that annotators consider the closer comparison image to the reference image as having the similar object as a function of triplet distance.         \textbf{CLIP and DINOv2 best reflect annotator perceptions of object similarity across all locations.}}
        \label{fig:triplet_distance}
\end{figure}

We next study the alignment between annotator perception of similarity and relative distances between reference and comparison images using different feature extractors,\footnote{We correct for the bias of using InceptionV3 distances when selecting images for annotations by subsampling to triplet distances for which there exist samples for all feature extractors. We normalize the plot by the maximum triplet distance and rescale the values to compare different feature extractors.} with results shown in Figure~\ref{fig:triplet_distance}.
We consider a good feature extractor one that has a high rate of alignment between the comparison image that is closer to the reference image in the feature space and the image that annotators consider more similar across all triplet distances, indicating that it correlates well with human judgment.

For all models, the chance that annotators select the closer comparison image as the more similar one is close to random for low triplet distance values. 
However, the probability of alignment grows as the triplet distance increases, and annotators almost always identify the image that is closer to the reference image as the more similar one at large distances.  
Both DINOv2 and CLIP features correlate significantly better with human judgement than InceptionV3, with DINOv2 features are slightly more associated with human judgment than CLIP features at low distance values. 
This is true for all image regions and annotator locations, suggesting there is little effect of geography in assessing similarity.

\vspace{1mm}
\noindent\fbox{    \parbox{0.98\textwidth}{        \textbf{Recommendation}: Use modern feature extractors like CLIP and DINOv2 for automatic measures of object similarity.
    }}

\subsection{Visual Appeal}

\subsubsection{Human Interpretations}

\begin{figure}
     \centering
     \begin{subfigure}[b]{0.35\textwidth}
             \includegraphics[width=0.99\textwidth]{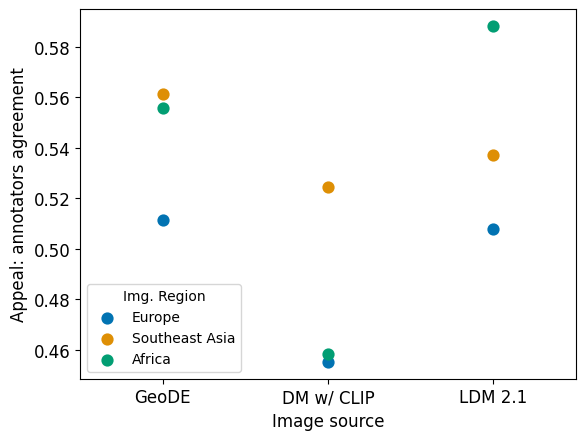}
            \vspace{2mm}
            \caption{Agreement on visual appeal}
            \label{fig:agreement_appeal_a}
            \end{subfigure} 
          \begin{subfigure}[b]{0.35\textwidth}\includegraphics[width=0.99\textwidth]{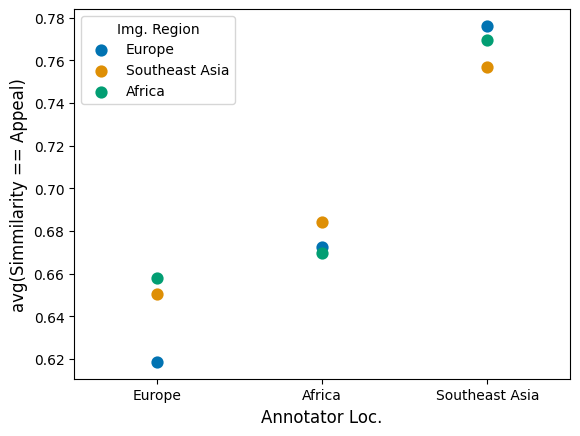}
             \vspace{2mm}
             \caption{Relationship b/w appeal \& similarity}
             \label{fig:agreement_appeal_b}
     \end{subfigure}
     \begin{subfigure}[b]{0.29\textwidth}
         \includegraphics[width=0.8\textwidth]{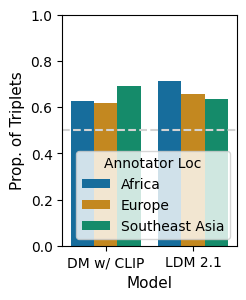}
                  \caption{Appealing img in real manifold}
         \label{fig:rep_human_auto_quant_aa}
     \end{subfigure}
        \vspace{-3mm}
        \caption{\textbf{(a)} Annotator agreement in visual appeal. 
        \textbf{(b)} Co-occurrence of appeal and similarity. 
        \textbf{(c)} Proportion of generated appealing images that fall in the GeoDE manifold.
        \textbf{Annotators agree on appeal less than similarity and at different rates between depicted regions. They may leverage qualities for appeal beyond similarity more when they have greater familiarity with what is in the image.
        The majority of appealing annotations of generated images fall in the manifold of real images.}}
                        \label{fig:appeal}
\end{figure}

To understand annotator subjectivity in determining visual appeal, we measure annotator agreement in deciding which image looks more appealing to them (Figure~\ref{fig:agreement_appeal_a}). 
We observe that the agreement is the lowest for images depicting Europe. Furthermore, annotators tend to agree on which image is more appealing more often for GeoDE and \ldmacc than \dmacc. 
In the Appendix, Figures~\ref{fig:qualitative_agreement_appeal} and~\ref{fig:qualitative_no_agreement_appeal} show examples where annotators agree and disagree, respectively, that one images is more appealing. 
We find that consistent appeal can occur when one sample has significantly lower visual quality. 
However, agreement is lower when annotators are shown two high quality images. 

Moreover, annotators in different locations tend to agree more on similarity than appeal, suggesting perceptions of appeal are more variable across regions.
This could be because similarity is grounded in a reference image, while visual appeal evokes latent interpretations of attractiveness. 
In addition, Figure~\ref{fig:agreement_appeal_b} shows that annotators in Southeast Asia more often select the same image as more similar \textit{and} appealing, suggesting that their interpretation of appeal coincides with that of similarity  moreso than annotators in other regions. 
For annotators across all locations, appeal coincides with similarity the least for images that correspond to their  location, suggesting that annotators may leverage qualities for appeal beyond similarity more when they have greater familiarity with what is presented in the image.

\vspace{1mm}
\noindent\fbox{    \parbox{0.98\textwidth}{        \textbf{Recommendation}: Collect annotations of visual appeal from annotators in multiple regions.
        While collecting more preferences will likely not resolve disagreements in aesthetic preferences, they may suggest directions for more inclusive evaluation criteria or definitions. Furthermore, they may be useful in challenging a universal notion of appeal by pinpointing specific preferences that vary between certain groups. 
    }}

\subsubsection{Human \& Automatic Metric Interaction: Manifold Presence}

Because visual appeal is approximated with precision, we construct an analogue to precision by counting the proportion of visually appealing annotations for generated images that fall within the manifold of all the real images in GeoDE.
In this set-up, GeoDE is representative of the real world's geographic diversity.
Figure~\ref{fig:rep_human_auto_quant_aa} shows that for most objects and regions, the majority of visually appealing annotations of generated images fall within the manifold of real images.

In Appendix Figure~\ref{fig:manifold_qual_close}, we inspect examples and find that annotators often pick images similar to those in GeoDE are more appealing, \textit{i.e.}, realistic, with the object in at least 25\% of the image, and high quality \citep{ramaswamy2022geode}. 
This tends to occur more for images that fall within the manifold, although some appealing images with these characteristics are outside the manifold. 
However, we also find that some annotators' interpretation of ``visually appealing'' diverge from GeoDE and fall outside of the manifold. 
Examples are shown in Appendix Figure ~\ref{fig:manifold_qual_far}.
In some cases, annotators give greater weight to the aesthetics of the background than the object. 
This is especially true when images have greater background stereotyping, such as for images of \textit{cars} in \ldmacc. Annotators can also prefer realistic objects even if they are lower quality or smaller. 
For example, a far-away dog is marked as more appealing than a close-up dog with unrealistic colors. 
The ``niceness'' of objects also relates to visual appeal as annotators tend to consider dirty, rusty cars, dull pots, and plain bags less appealing than newer, more pristine, and colorful counterparts. 

\vspace{1mm}
\noindent\fbox{    \parbox{0.98\textwidth}{        \textbf{Recommendation}: When using manifold-based metrics to measure visual appeal, select a reference dataset that aligns with the definition used by annotators or needed for model deployment. When reporting on appeal, clarify whose and what definition is captured, including whether appeal is associated with exaggerations or stereotypes.
    }}

\subsection{Object Consistency}

\begin{figure}
    \centering
    \begin{subfigure}[b]{0.22\textwidth}
         \includegraphics[width=0.95\textwidth]{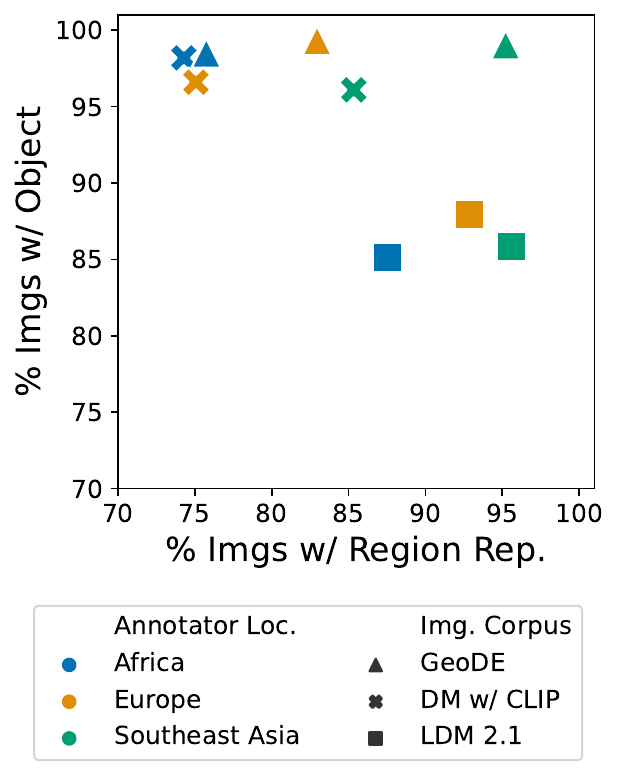}
         \vspace{6mm}
         \caption{Geographic representation vs. object appearance}
         \label{fig:consistency_human_quant}
    \end{subfigure}
    \hspace{2mm}
    \begin{subfigure}[b]{0.38\textwidth}
         \includegraphics[width=0.95\textwidth]{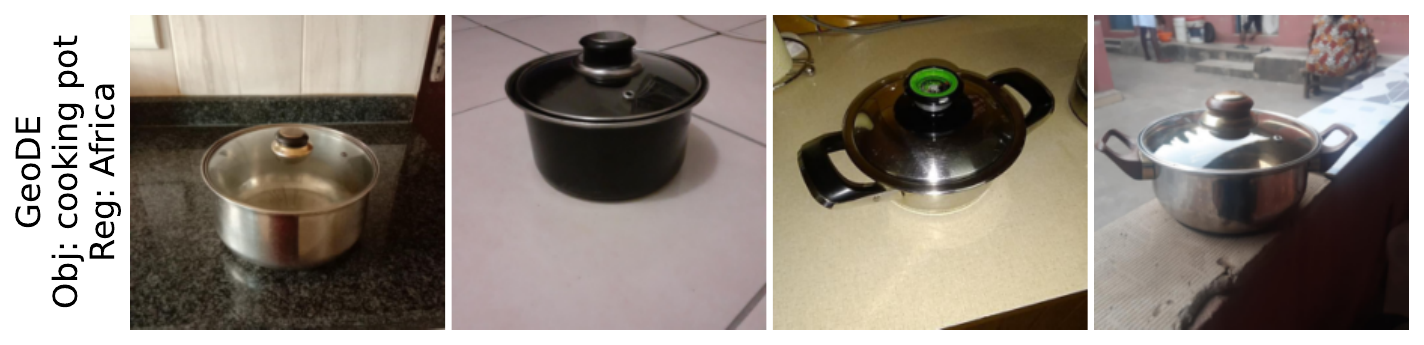}
         \includegraphics[width=0.95\textwidth]{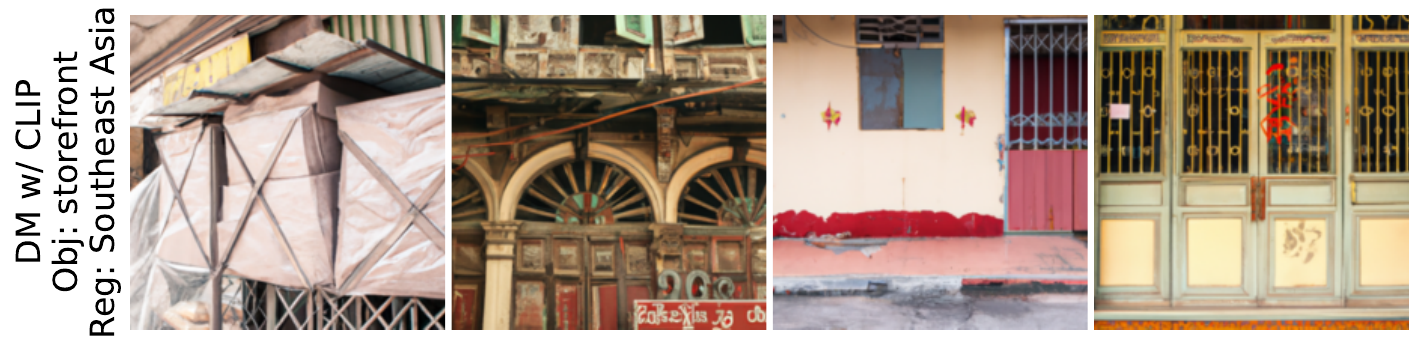}
         \includegraphics[width=0.95\textwidth]{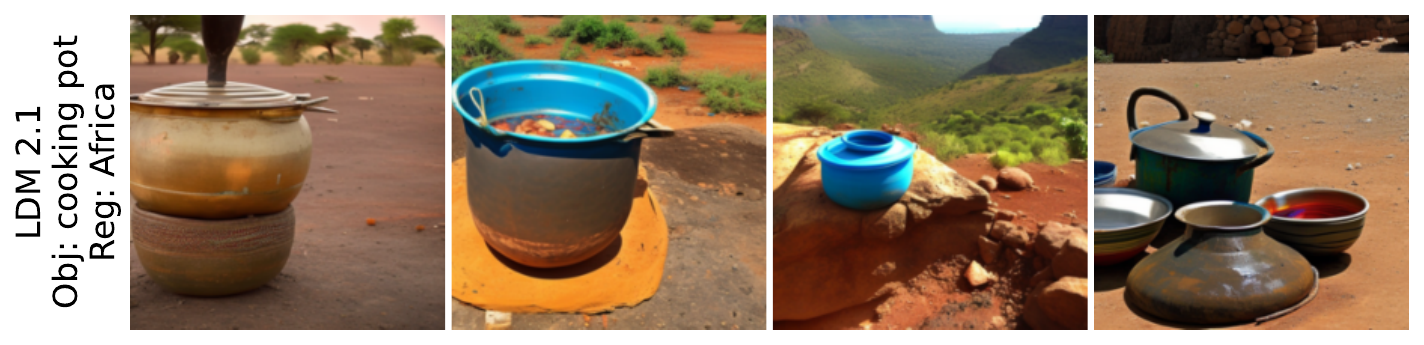}
         \vspace{7mm}
         \caption{Examples where in-region ann. says object is not shown and out-of-region ann. disagrees.}
    \label{fig:consistency_human_qual}
     \end{subfigure}
    \hspace{2mm}
    \begin{subfigure}[b]{0.34\textwidth}
         \includegraphics[width=0.95\textwidth]{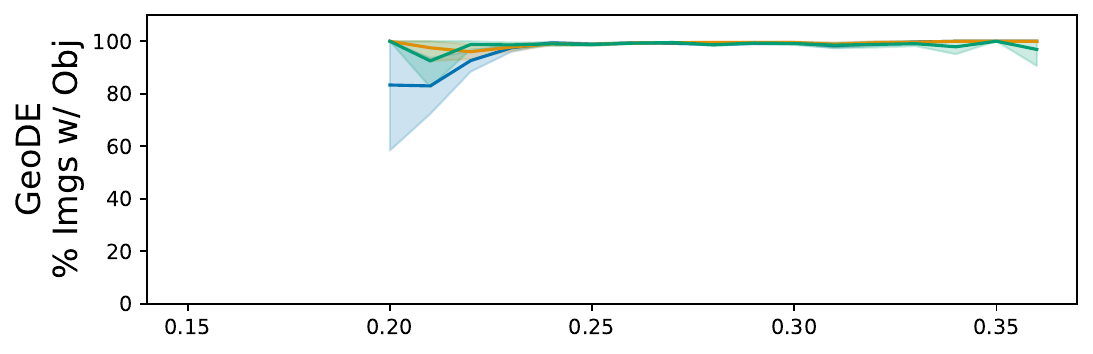}
         \includegraphics[width=0.95\textwidth]{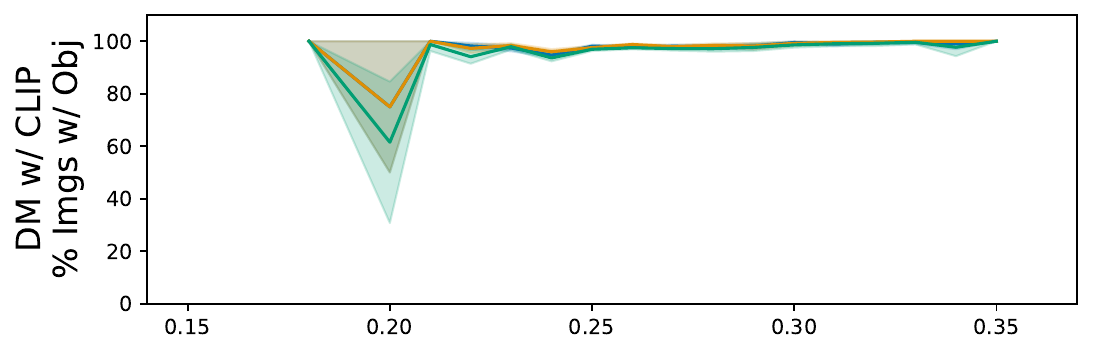}
         \includegraphics[width=0.95\textwidth]{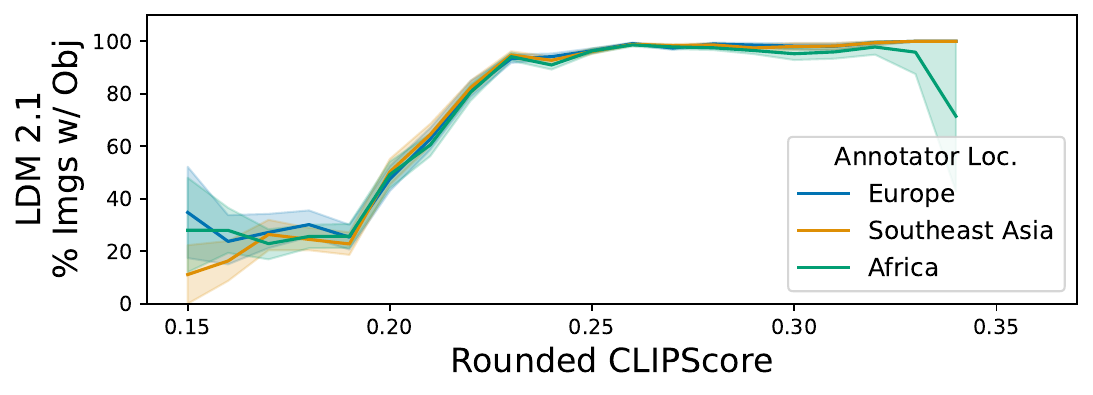}
    \caption{Relationship between Object-CLIPScore and presence of object in image.}
    
    \label{fig:rep_human_auto_quant}
    \end{subfigure}
    \caption{\textbf{(a)} Relationship between object consistency and \georep.
    \textbf{(b)} Disagreements in consistency between annotators.
    \textbf{(c)} Comparison of Object-CLIPScore and annotator perception of object consistency.
    \textbf{Annotators in different regions vary slightly in their recognition of concrete objects. Beyond a threshold, Object-CLIPScore corresponds to consistent presence of objects for all annotator locations.}}
    \label{fig:consistency_quant}
\end{figure}

\subsubsection{Human Interpretations}
Finally, we investigate the rate at which annotators identify the desired object in a photo and the extent to which this coincides with the \georep of the image.
In  Figure ~\ref{fig:consistency_human_quant}, we find that annotators across all locations consider \ldmacc as having the least object consistency while also indicating that \ldmacc tends to have the greatest \georep. 
This aligns with qualitative observations from~\citep{hall2023dig} that \ldmacc often depicts strong \georep at the cost of object consistency. 

In Figure~\ref{fig:consistency_human_qual}, we visualize examples in which in-region annotators say the desired object is not shown while out-of-region annotators say it is. 
For example, annotators located in Southeast Asia do not consider a subset of images from \dmacc generated with the prompt ``\textit{storefront in Southeast Asia}'' as storefronts while annotators located outside of Southeast Asia do. 
The same pattern is true of in-region and out-of-region annotators for generated images of a ``\textit{cooking pot in Africa}.'' 
This suggests that annotators located in different regions vary in what they consider to be accurate visual representations of concrete objects. 
Interestingly, in-region annotators can even disagree about objects in real images that were collected from people living in that region, as in the example of cooking pots from GeoDE in the Figure. 
This suggests possible \textit{intra}-region variations in the definition of objects as well as \textit{inter}-region variations.

\vspace{1mm}
\noindent\fbox{    \parbox{0.98\textwidth}{        \textbf{Recommendation}: When annotating for object consistency, include perspectives from in- and out-of-region. Multiple annotators per region are ideal for capturing within-region variations. If uniformity is desired, instruct annotators to focus on a specific object definition / representation, including positive and negative examples.
    }}

\subsubsection{Human \& Automatic Metric Interaction: Object-CLIPScore}

In Figure~\ref{fig:rep_human_auto_quant}, we plot bucketed CLIPScores and average per-image annotations of whether the object appears in the image. 
By examining \ldmacc, which suffers from some consistency challenges, we find that the lower range of CLIPScores is a useful indicator of whether the object appear in the image.
This is aligned with findings in ~\citep{hall2023dig}.
However, at higher scores the object is consistently in the image. 
As with the Region-CLIPScore, we see that the range of CLIPScores can vary between types of images. 
We find empirically that a threshold of 0.21-0.25 best indicates consistent appearance of an object. 
This range accounts for  variations between datasets, with a threshold ~0.21 corresponding to nearly constant object presence for GeoDE and \dmacc and ~0.25 for \ldmacc. 
Unlike the Region-CLIPScore, the Object-CLIPScore seems equally informative regardless of which location the annotator is in. 

\vspace{1mm}
\noindent\fbox{    \parbox{0.98\textwidth}{        \textbf{Recommendation}: Use a threshold for Object-CLIPScore to include multiple representations of the same concept.
    }}
\section{Related Work}
\label{sec:related_work}

Beyond text-to-image models, our work is situated within a wider discussion of geographic and cultural performance disparities in computer vision systems. 
For example, \citet{devries2019does} report that object detection models suffer marked drops in performance when evaluated on images from non-Western geographies, and particularly so for those from lower-income countries.
\citet{goyalfairness2022} find similar disparities in the performance of visual feature extractors, as does \citet{gustafson2023pinpointing} for the popular image-text foundation model CLIP \citep{radford2021learning}. 
These failures are likely to be, at least in part, the result of poor representation in the training datasets: ImageNet \citep{deng2009imagenet} displays a sample that is overwhelmingly biased toward North American and Western European contexts \citep{shankar2017no}.
Remarkably, \citet{richards2023progress} report that as performance has improved on classic object detection benchmarks, the performance gap \emph{between} regions has worsened.~\looseness-1

Accounting for geographic context is a challenge not limited to vision systems: in NLP, a growing body of research investigates subjectivity among annotators \cite{al-kuwatly-etal-2020-identifying,goyal2022toxicity,DBLP:journals/corr/abs-2110-05719} and, in particular, cross-cultural sensitivity.
While cultural diversity is not exactly proxied by geographic diversity, we expect cultural differences to often co-occur with geographic shifts, and consider work on cross-cultural inclusion deeply relevant to our work.
Researchers working on machine translation have long sought to improve translation both to and from so-called low-resource languages,
though models continue to exhibit failures along culturally-specific axes \citep{akinade2023}. Recent work explores the role of cultural context in automated detection of stereotypes \citep{jha2023, dev2023building}, detecting toxic text or hate speech \citep{lee2023hate}, and values espoused by large language models \citep{arora2023probing, durmus2023subjective}.
Others have focused on how evaluation practices suffer from a narrow cultural frame. 
\citet{hutchinson2022gaps} explore how cultural factors are overlooked during NLP system evaluation, highlighting the obfuscation of annotators' ``socio-demographic standpoints'' [p.~1867].
Similarly, \citet{prabhakaran2022incongruencies} call for ``culturally situated'' evaluations [p.~3], while \citet{sambasivan2021fairness} argue that fairness evaluations must account for non-Western perspectives on fairness itself.
Our work responds to such calls, exploring the extent to which automated evaluation metrics can account for geographic differences and cultural expectations. \looseness=-1
\section{Conclusion}
\label{sec:conclusion}
We study how annotators in different regions vary in their perceptions of evaluation criteria for text-to-image generative models and how well automatic metrics capture these variations. 
From our analyses, we recommend that annotations of \georep include perspectives from people located inside \textit{and} outside regions of focus and that instructions clarify whether subjectivity and reference to external sources are encouraged.
In addition, modern feature extractors like CLIP and DINOv2 can better capture aspects of image similarity when performing automatic evaluations than the standard Inceptionv3.
We find that visual appeal is often interpreted in contradictory preferences that can reinforce stereotypes and recommend improving automatic metrics like precision via careful selection of reference datasets.
Perceptions of object consistency can have \textit{within} region variations but are generally well approximated by CLIPScore across all annotator locations. 
Finally, contradictory responses for criteria like appeal, similarity, and consistency can reveal true ambiguity that is lost in ``majority-vote'' type aggregations.
We hope that this work is a meaningful step towards greater geographic inclusion in human and automatic evaluations of text-to-image generative models.~\looseness-1

\vspace{1mm}
\textit{Limitations.}
Our study is limited to three broad regions, six objects, and three image sources.
Annotators may vary in their familiarity of objects presented in the study.
Due to the vast variations in human subjectivity within regions, findings based on annotator responses are limited in their extensibility to those not included in the study.
Task construction may also have affected annotator responses, and conclusions are limited without a deep focus group to collect qualitative feedback and explanation from annotators.
~\looseness-1

\begin{acks}
We thank Carleigh Wood, Ida Cheng, Emerson Bacud, and Adam Hakimi for their many contributions throughout the annotation process. 
We also thank Laura Gustafson, Mark Ibrahim, and Pietro Astolfi for providing technical advice.
\end{acks}

\newpage

\paragraph{Ethical considerations}
This work would not have been possible without dozens of individual human annotators. 
In our data collection process, we followed recommendations pertaining to responsible data collection, including (i) compensating annotators fairly for their work at an hourly rate, (ii) respecting the privacy of annotators by using anonymous ids in the data collection process, (iii) avoiding questions about personal information, (iv) providing voluntary and informed consent about the annotation task and use of data, (v) providing multiple bi-directional channels of communication with annotators, and (vi) allowing annotators to opt-out of the task at any time.
In addition, we attempted to mitigate risks of exposure to harmful or uncomfortable content by focusing on images of everyday objects rather than people or divisive topics.

\paragraph{Researcher positionality}
The authors of this work recognize the role that our personal experience and institutional resources and constraints play in the annotation collection process and subsequent analyses and recommendations. 
The choice of questions, objects, regions, and images included in the annotation tasks were selected with the intention of increasing the variety of perspectives about annotator perceptions of evaluation criteria within our infrastructure and financial constraints. 
While we attempted to mitigate bias in our interpretation of quantitative annotations by including a qualitative survey, not all annotators completed the survey. We therefore remain susceptible to our subjective interpretation of responses. 
Furthermore, it is likely that nuance in annotator interpretation of the evaluation criteria is lost in our quantification and summarization of analyses, even with our effort to inspect qualitative examples. 
In addition, the interpretation and coding of the qualitative survey data was conducted by three authors based in Europe or North America. 
Our inductive coding approach is likely to be biased by our own experiences.
Finally, our recommendations pre-suppose a degree of resourcing that is not accessible to all, such as the potentially costly suggestion of collecting multiple annotations for a single sample to capture subjectivity. 
While we hope this work informs improved evaluations of text-to-image generative models, there is much room for further investigation of the metrics and annotator behaviors.

\paragraph{Adverse impacts}
Findings about variations in human and automatic evaluations studied in this work may differ when applied to other types of images, such as depictions of people, complex scenes, or creative imagery, or across different demographic groups such as gender or age. 
Readers should use caution when extending findings to broader regional trends or when informing future annotator patterns.
A future study would be strengthened by including multiple annotators per region for the same task, to account for intra-region variability. 

\newpage

\bibliographystyle{ACM-Reference-Format}
\bibliography{sample-base}

\appendix
\section{Appendix}
\label{sec:appendix}

We discuss additional details regarding the set-up and results of our analysis in the following sections. 

\subsection{Additional Set-up Details}
\label{app:set_up}

\paragraph{Tasks}

For Task 1, we construct triplets with real images from GeoDE and generated images from \dmacc and \ldmacc. 
In order to have representation across different types of real and generated images as well as relative distances in the embedding space, we define different combinations of triplets each object-region as follows:

\begin{itemize}
    \item[A] One real image and two generated images, for 85 real images of object $O$ in region $R$. 
    \begin{itemize}
        \item Reference Image: One real image of $O$ from $R$.
        \item Comparison Images: One image generated with prompt ``$O$ in $R$'' which is the closest to the reference image in the feature space. and one image generated with prompt ``$O$ in $R$'' which is the eighth furthest from the reference image.
    \end{itemize}

    \item[B] One generated image and two real images, for 85 images generated with prompt ``$O$ in $R$''.
    \begin{itemize}
        \item Reference Image: One image generated with prompt ``$O$ in $R$''.
        \item Comparison Images: One real image of $O$ from $R$ which is the closest to the reference image in the feature space, and one real image of $O$ from $R$ which is the eighth furthest from the reference image.
    \end{itemize}

    \item[C] One real image and two generated images, for 85 real images of object $O$ in region $R$.
    \begin{itemize}
        \item Reference Image: One real image of $O$ from $R$.
        \item Comparison Images: One image generated with prompt ``$O$ in $R$'' which is the closest feature to the reference image, and one image generated with the prompt ``$O$'' that is the 48th furthest from the reference image.
    \end{itemize}

    \item[D] One real image and two generated images, for 85 real images of object $O$ in region $R$.
    \begin{itemize}
        \item Reference Image: One real image of $O$ from $R$.
        \item  Comparison Images: One image generated with prompt ``$O$'' which is the sixth feature to the reference image, and one image generated with the prompt ``$O$ in $R$'' that is the eighth furthest from the reference image.
    \end{itemize}

    \item[E] One generated image and two real images, for 85 images generated with prompt ``$O$''.
    \begin{itemize}
        \item Reference Image: One image generated with prompt ``$O$''.
        \item Comparison Images: One real image of $O$ from $R$ which is the closest to the reference image in the feature space, and one real image of $O$ from $R$ which is the eighth furthest from the reference image.
    \end{itemize}
\end{itemize}

To select reference images, we pick the 60 images for each object-region combination that has another image nearest to it and the 25 images that has its nearest image furthest away. 
This allows us to capture statistics about both images that are relatively similar to the others (the former) and images that are quite different (the latter).

For our Visual Appealing analysis, we use Combo A.
This allows a comparison between real images of a given object taken in a specific region and generated images prompted with the same object and region.
In addition, for the Visual Appeal analysis we select only triplets where exactly one image falls into the GeoDE manifold.

\paragraph{Annotators}

Potential annotators were first recruited according to their geographic location.
They were provided an initial task to complete to demonstrate general competency then went through a training program consisting of written instructions for each task and practice examples. 
Because our Tasks require an amount of subjectivity throughout them yet we still want to ensure annotators have a clear understanding of the intended task, we created a trial task that involved annotators identifying similar images of cars. 
Annotators who passed with a 75\% rate were then graduated to the production tasks and paid a standard rate per hour based on a country-specific living wage. 

For production tasks, annotators within a given location worked through queues for Task 1 and Task 2 for each object and image region. 
They varied in the proportion of jobs completed, with some annotators completing less than 10 jobs and others hundreds jobs. 
Each unique job was annotated by one annotator from each region.

Following the completion of the production tasks, we performed an analysis of annotation completion quality.
We use annotations of whether the real GeoDE images contain a given object as our ``sanity'' check, as we have a ground truth answer for these images.
We found that while the accuracy for real GeoDE images was quite high for most annotators, a very small number of annotators had low accuracy for identifying objects in GeoDE that were included as comparison image in the task construction.
In those cases, we believe the annotators did not fully understand the task instructions to select \textit{all} images showing the object or were operationalizing the task incorrectly. 
Thus, we introduce a requirement that annotators have at least 90\% accuracy in identifying objects in GeoDE images, to ensure some conformity to definition of object representation and task completion. 
For each object-region combination, we remove annotations from annotators who did not meet this bar. 

At the conclusion of filtering, we had annotations from 20 annotators located in Africa (split across Egypt, Nigeria, and South Africa), 17 annotators in Europe (located in Great Britain, Italy, Romania and Spain) and 23 annotators in Southeast Asia (located in Indonesia, the Philippines, and Thailand). 
All annotators were fluent in English, and tasks were presented in English.

\subsection{Additional annotation details}

\begin{figure}
    \centering
    \begin{subfigure}[b]{0.64\textwidth}
        \includegraphics[trim={0 0 30cm 0},clip,width=1.0\textwidth]{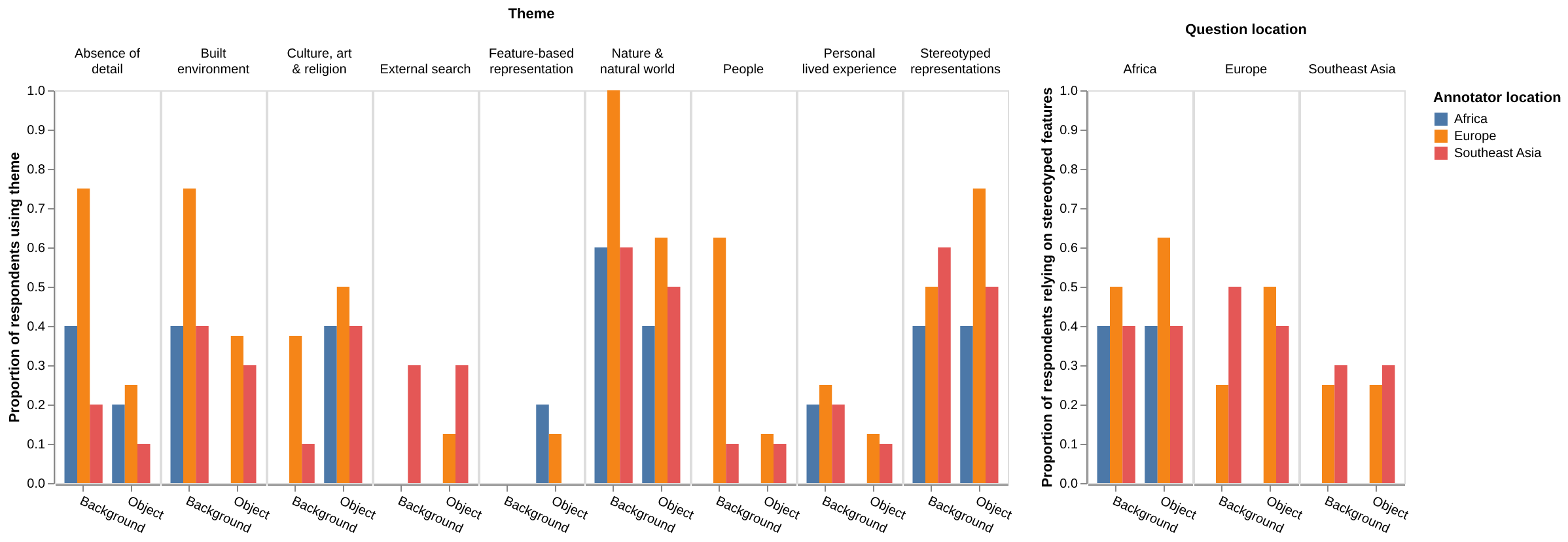}
                        \caption{}
        \label{fig:rep_human_survey_themes}
    \end{subfigure}
            \begin{subfigure}[b]{0.35\textwidth}
        \includegraphics[trim={55cm 0 0 0},clip,width=1.0\textwidth]{annotation_theme_and_stereotyping_combined.png}
        \caption{}
        \label{fig:rep_human_survey_stereotyping}
    \end{subfigure}
    \caption{\textbf{(a)} Proportion of survey respondents exhibiting key themes in their responses to questions about identifying locations from both background context and object specifics. \textbf{(b)} Proportion of survey respondents suggesting they utilized common stereotypes when annotating geographic representativeness. \textbf{Annotators in all locations frequently exhibit stereotypes in their responses.}}
    \label{fig:rep_human_survey_all}
\end{figure}

Table \ref{table:survey-codes} contains descriptive codes used to categorize responses from annotators. Figure \ref{fig:rep_human_survey_all} shows proportion of survey respondents exhibiting key themes in their response to questions about their interpretation of the annotation task.

\begin{table*}
\footnotesize
\centering

\begin{tabular}{llr}
\toprule
                           Code &                                                             Description &  Kappa \\
\midrule
                External search &                                   Using an external reference, e.g. search engine &   0.91 \\
                         People &                                                      Mentions people or faces &   0.80 \\
      Nature \& natural world & Mentions nature, plants, animals, landscape, weather or climate &   0.77 \\
Feature-based representation &       Mentions looking for feature-based representations, e.g. looking at color, size or shape &   0.76 \\
                   Unclassified &                                                  Unclassified response &   0.73 \\
 Stereotyped representations &       Utilizes common stereotypes of appearance of items within a region &   0.72 \\
           Built environment &                      Mentions buildings, architecture, or the built environment &   0.71 \\
           Absence of detail &                            Mentions an absence detail or noteworthy characteristics &   0.70 \\
     Culture, art \& religion &         Mentions culture, food, art, sculpture, religion, fashion and clothing &   0.63 \\
 \midrule
   Personal lived experience &                                Leverages experience living in a location &   0.60 \\
               Media portrayals &                                         Reliance on portrayals in media &   0.60 \\
           Brand identification &                   Uses a known reference, such as recognising a brand or store &   0.52 \\
Overgeneralised representations &            Considers subsets of a region indicative of a broader entire region &   0.48 \\
                Objects and tools &                      Uses everyday objects such as tools or appliances &   0.41 \\
       Erroneous representation &            Mentions erroneous or unrealistic object or background representation &   0.41 \\
                     Uniqueness &                     Expectation that a representation be unique to a region &   0.15 \\
\bottomrule
\end{tabular}
\caption{Descriptive codes used to categorize survey responses from annotators. Codes are sorted by inter-annotator agreement Fleiss' Kappa, and only codes with Kappa greater than 0.6 (horizontal rule) were included in our analysis.}
\label{table:survey-codes}
\end{table*}

\subsection{Additional Results}
\label{app:additional_results}

\paragraph{Visual appeal}

\begin{figure}
     \centering
     \begin{subfigure}[b]{0.30\textwidth}
         \centering
         \includegraphics[width=\textwidth]{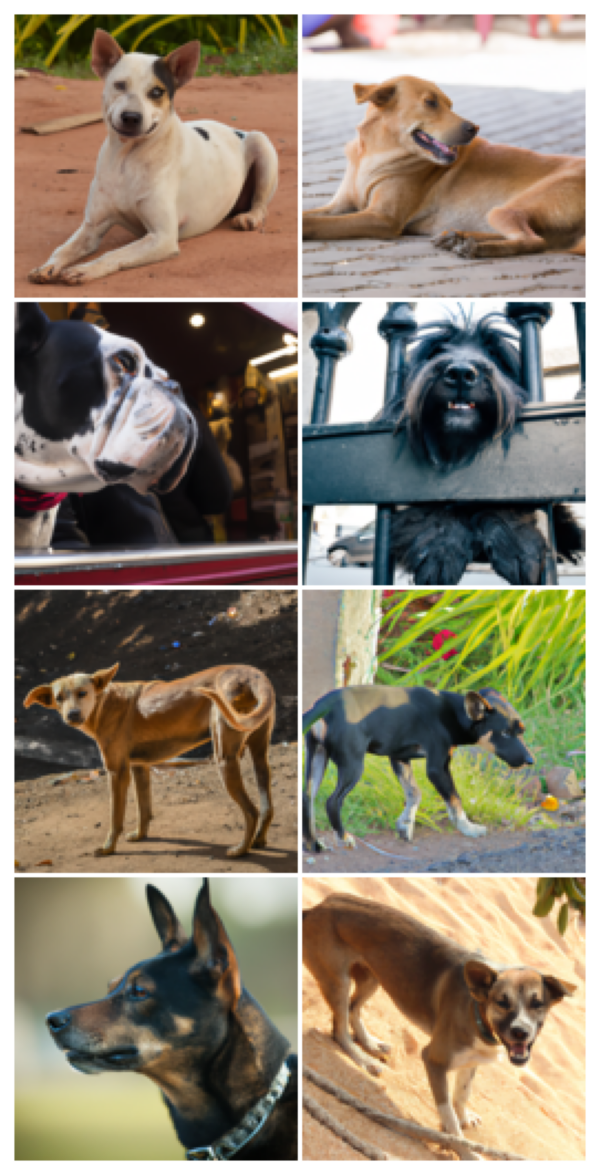}
         \caption{Consistent appeal}
     \label{fig:qualitative_agreement_appeal}
     \end{subfigure}
           \begin{subfigure}[b]{0.30\textwidth}
         \centering
    \includegraphics[width=\textwidth]{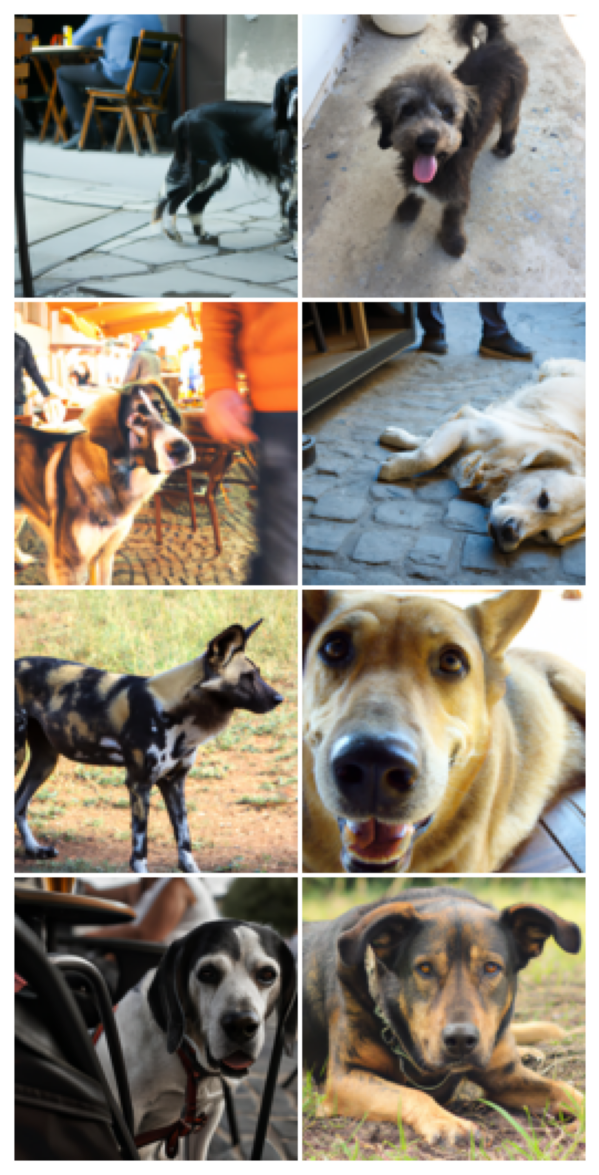}
         \caption{Inconsistent appeal}
    \label{fig:qualitative_no_agreement_appeal}
     \end{subfigure}
        \vspace{-3mm}
        \caption{
        \textbf{(a-b)} Examples of generated images of \textit{dogs}: (b) depicts images selected as more and less appealing (respectively) by all three annotators; (c) depicts images with inconsistent appeal annotations.
        \textbf{Annotators may leverage qualities for appeal beyond similarity more when they have greater familiarity with what is in the image.}}
                \label{fig:appeal_app}
\end{figure}

\begin{figure}
    \centering
         
    \begin{subfigure}[b]{0.48\textwidth}
\includegraphics[width=.99\textwidth]{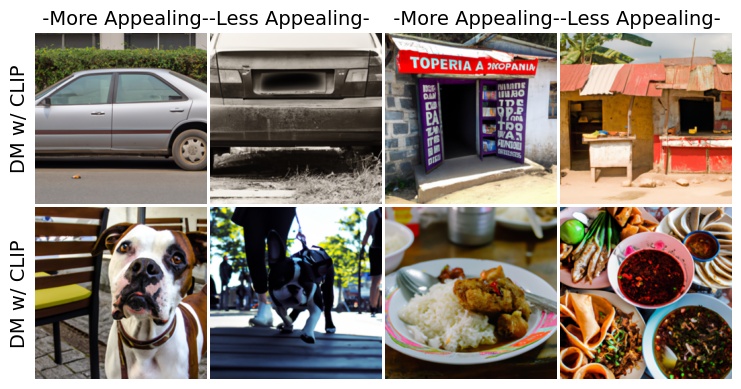}
        \includegraphics[width=.99\textwidth]{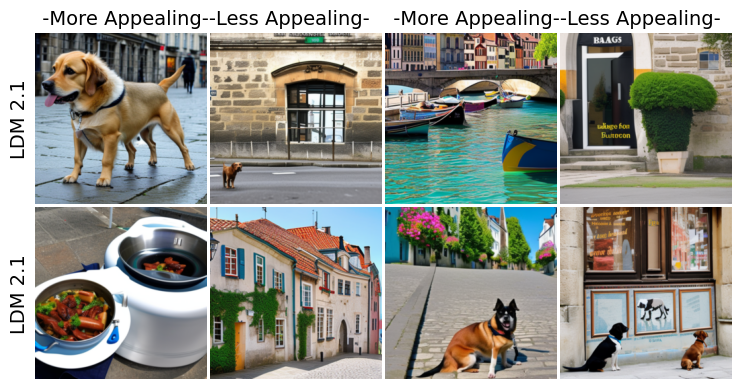}
        \caption{Examples where \textit{more} appealing generated image falls in GeoDE manifold}
        \label{fig:manifold_qual_close}
     \end{subfigure}
     \hspace{1mm}
         \begin{subfigure}[b]{0.48\textwidth}
        \includegraphics[width=.99\textwidth]{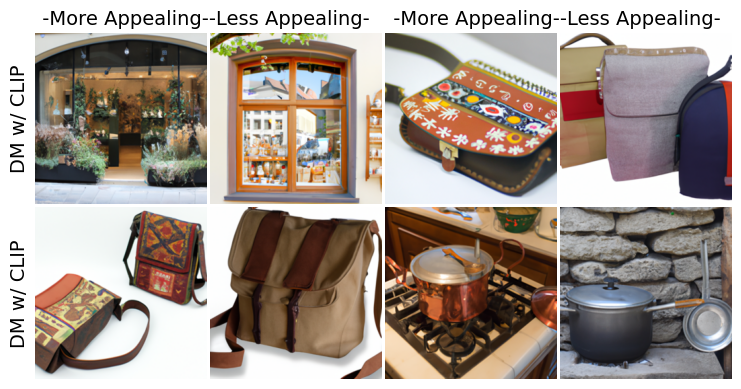}
        \includegraphics[width=.99\textwidth]{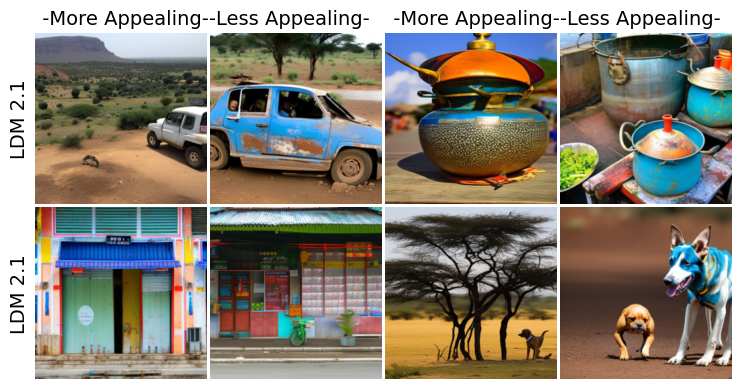}
        \caption{Examples where \textit{less} appealing generated image falls in  GeoDE manifold}
        \label{fig:manifold_qual_far}
     \end{subfigure}
     \vspace{-2mm}
    \caption{ \textbf{(a-b)} Example pairs where the \textit{more} or \textit{less} appealing image falls in the manifold of real images.
    \textbf{
    Annotators often consider realistic images more appealing, aligning with the real manifold, but also interpret appeal in contradictory ways, \textit{e.g.}, prioritizing background aesthetics, stereotypes, or ``niceness'' of objects.}}
    \label{fig:rep_human_auto_appeal_app}
\end{figure}

\begin{figure}
     \centering
     \includegraphics[width=\textwidth]{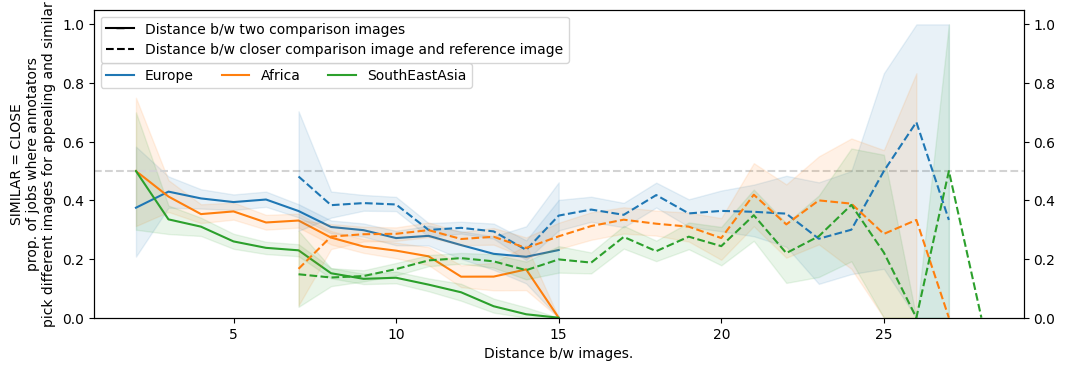}
     \includegraphics[width=\textwidth]{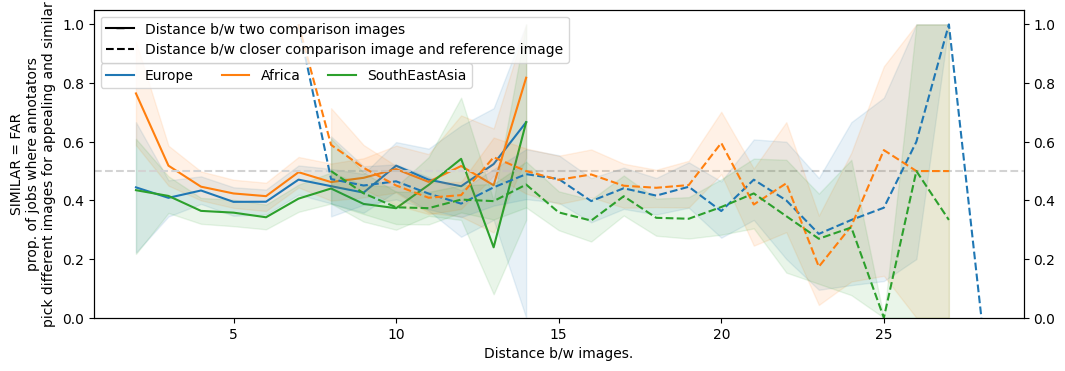}
     \caption{When annotators pick the further reference image as the more similar image, all annotators differentiate \textit{more} between similarly and visually appealing than when they were picking the closer image as the similar image. }
     \label{fig:visual_appeal_differentiate}
\end{figure}

The top of Figure \ref{fig:visual_appeal_differentiate} shows that when annotators pick the comparison image that is closer to the reference image as the more similar image, all annotators differentiate between visual appeal and similarity \textit{less} as comparison images move away from each other in InceptionV3 space (solid line) and \textit{slightly more} as images move further from the reference image (dashed line).
Furthermore, annotators in Europe tend to differentiate between similar and visually appealing images \textit{more} than annotators in Africa, and both more than annotators in Southeast Asia. 
This is consistent even as distances between comparison and reference images vary. 

In addition, the bottom of Figure \ref{fig:visual_appeal_differentiate} shows that when annotators pick the further image as the more similar image, all annotators differentiate \textit{more} between similarly and visually appealing than when they were picking the closer image as the similar image. 
However, this differentiation seems less impacted by the distance of comparison and reference images, and there is less between-region variation.

\end{document}